\pdfoutput=1
\documentclass[11pt]{article}

\usepackage[final]{acl}

\usepackage{times}
\usepackage{latexsym}

\usepackage[T1]{fontenc}
\usepackage[utf8]{inputenc}

\usepackage{microtype}

\usepackage{inconsolata}

\usepackage{graphicx}
\usepackage{url}
\usepackage{latexsym}
\usepackage{makecell}
\usepackage{epstopdf}
\usepackage{tipa}
\usepackage{hyperref}
\usepackage{xstring}
\usepackage{amsfonts}
\usepackage{amsmath}
\usepackage{fixltx2e}
\usepackage[T1]{fontenc}
\usepackage{booktabs}
\usepackage{arabtex}
\usepackage{hhline}
\usepackage{pdfpages}
\usepackage{utf8}
\newcommand{\hide}[1]{}

\newcommand{\AHAMZAUP}{{\^{A}}}

\newcommand{\AHAMZADN}{{\v{A}}}

\newcommand{\AMAQ}{{\'{y}}}

\usepackage{multirow}
\usepackage{subfigure}
\usepackage{amssymb}
\usepackage[super]{nth}
\usepackage{arydshln}
\usepackage{bbding}
\usepackage[framemethod=TikZ]{mdframed}
\usepackage{caption}
\novocalize
\usepackage{xcolor}

\title{Enhancing Text Editing for Grammatical Error Correction: \\ Arabic as a Case Study}

\author{First Author \\
  Affiliation / Address line 1 \\
  Affiliation / Address line 2 \\
  Affiliation / Address line 3 \\
  \texttt{email@domain} \\\And
  Second Author \\
  Affiliation / Address line 1 \\
  Affiliation / Address line 2 \\
  Affiliation / Address line 3 \\
  \texttt{email@domain} \\}

\author{Bashar Alhafni\textsuperscript{$\dagger$} \and Nizar Habash\\
  Computational Approaches to Modeling Language Lab\\
  New York University Abu Dhabi\\
  \textsuperscript{$\dagger$}Mohamed bin Zayed University of Artificial Intelligence\\
  \texttt{\{alhafni,nizar.habash\}@nyu.edu}
  }
  
\begin{document}
\setcode{utf8}
\setarab
\maketitle
\begin{abstract}
Text editing frames grammatical error correction (GEC) as a sequence tagging problem, where edit tags are assigned to input tokens, and applying these edits results in the corrected text. This approach has gained attention for its efficiency and interpretability. However, while extensively explored for English, text editing remains largely underexplored for morphologically rich languages like Arabic. In this paper, we introduce a text editing approach that derives edit tags directly from data, eliminating the need for language-specific edits. We demonstrate its effectiveness on Arabic, a diglossic and morphologically rich language, and investigate the impact of different edit representations on model performance. Our approach achieves SOTA results on two Arabic GEC benchmarks and performs on par with SOTA on two others. Additionally, our models are over six times faster than existing Arabic GEC systems, making our approach more practical for real-world applications. Finally, we explore ensemble models, demonstrating how combining different models leads to further performance improvements. We make our code, data, and pretrained models publicly available.\footnote{\url{https://github.com/CAMeL-Lab/text-editing}}

% Text editing approaches frame grammatical error correction (GEC) as a sequence tagging problem, where edit tags are assigned to tokens in the input, and applying these edits results in the corrected text. This approach has garnered attention for its efficiency and interpretability. However, while it has been extensively explored for English, text editing has remained largely underexplored for morphologically rich languages like Arabic. In this paper, we introduce a novel text editing approach that derives edit tags directly from data, eliminating the need for linguistic knowledge. We demonstrate the effectiveness of our approach on Arabic, a diglossic and morphologically complex language, and investigate the impact of different edit representations on model performance. Our method achieves state-of-the-art results on two Arabic GEC benchmarks and matches current SOTA performance on two others. Furthermore, our model is more than six times faster than existing Arabic GEC systems, making it significantly more practical for real-world applications. Finally, we explore the potential of ensembling techniques, demonstrating how combining different models can lead to further performance improvements.

\end{list}
\end{abstract}

\section{Introduction}
Grammatical Error Correction (GEC) is a well-studied problem, particularly in English, with numerous datasets and shared tasks \cite{ng-etal-2013-conll,ng-etal-2014-conll,bryant-etal-2019-bea}. GEC has applications in both writing assistance for native speakers (L1) and language learning for second-language (L2) learners. While neural machine translation (NMT) approaches have long dominated GEC and continue to achieve strong results when trained on large amounts of data \cite{stahlberg-kumar-2024-synthetic,bryant-etal-2023-grammatical}, they are not inherently the most efficient. Unlike MT, where input and output sequences differ significantly, GEC typically involves minimal changes, with most input tokens copied to the output. Employing full-sequence autoregressive models in such cases can be computationally wasteful \cite{stahlberg-kumar-2020-seq2edits}.

A highly efficient and competitive alternative to sequence-to-sequence (Seq2Seq) models is text editing, which frames GEC as a sequence tagging problem. Instead of generating text autoregressively, text editing models assign edit labels to input tokens, leading to a more efficient and interpretable corrections. However, most popular text editing approaches require effort to design language-specific edit tag sets \cite{awasthi-etal-2019-parallel, omelianchuk-etal-2020-gector, mesham-etal-2023-extended}. This limits their adaptability for morphologically rich languages like Arabic \cite{kwon-etal-2023-beyond}, where the space of possible edits is large.

Inspired by recent advancements in text editing \cite{awasthi-etal-2019-parallel,malmi-etal-2019-encode,omelianchuk-etal-2020-gector,straka-etal-2021-character,mesham-etal-2023-extended}, we introduce a novel text editing approach that eliminates the need for language-specific edits. Instead, our method derives edit tags directly from data, making it more adaptable and scalable across different linguistic settings. We demonstrate the effectiveness of our approach on Arabic GEC. Our contributions are as follows:

\begin{enumerate}
    \item We introduce the first successful application of text editing to Arabic GEC and study the effect of edit representation on the task.
    \item We achieve SOTA results on two Arabic GEC benchmarks and perform on par with SOTA on two others.
    \item Our models are over six times faster than existing Arabic GEC systems, making them more practical for real-world applications.
    \item We show through ensembling experiments how different models complement each other, leading to significant performance gains.
\end{enumerate}

\section{Background and Related Work}

\subsection{Grammatical Error Correction}
% \begin{itemize}
%     \item Seq2Seq approaches
%     \item Seq2Edit approaches and why they were intorduced
%     \item LLMs for GEC and how they still behind SOTA
% \end{itemize}

GEC has been approached using a variety of methods, with Transformer-based systems being the most popular \cite{bryant-etal-2023-grammatical}. The use of Transformer-based architectures in GEC began by framing the task as a neural machine translation (NMT) problem \cite{junczys-dowmunt-etal-2018-approaching,yuan-etal-2019-neural,zhao-etal-2019-improving,grundkiewicz-etal-2019-neural,katsumata-komachi-2020-stronger,kaneko-etal-2020-encoder,wan-etal-2020-improving,yuan-etal-2021-multi,yuan-bryant-2021-document,stahlberg-kumar-2021-synthetic,rothe-etal-2021-simple,zhou-etal-2023-improving-seq2seq,luhtaru-etal-2024-error}.

To improve efficiency and interpretability, text editing models have emerged as an alternative to Seq2Seq approaches \cite{awasthi-etal-2019-parallel,malmi-etal-2019-encode,stahlberg-kumar-2020-seq2edits,mallinson-etal-2020-felix,omelianchuk-etal-2020-gector,straka-etal-2021-character,mallinson-etal-2022-edit5,tarnavskyi-etal-2022-ensembling,mesham-etal-2023-extended,zhang-etal-2023-non}. Unlike Seq2Seq models, which generate corrected text from scratch, text editing models treat GEC as a sequence tagging task, producing a set of edit operations that modify the erroneous input. Our work follows this text editing paradigm.

LLMs have also been evaluated on GEC \cite{fang2023chatgpt,coyne2023analyzingperformancegpt35gpt4,wu2023chatgptgrammarlyevaluatingchatgpt,loem-etal-2023-exploring,raheja-etal-2023-coedit,kaneko-okazaki-2023-reducing,raheja-etal-2024-medit,davis-etal-2024-prompting,katinskaia-yangarber-2024-gpt,omelianchuk-etal-2024-pillars,mita-etal-2024-towards,kaneko-okazaki-2024-controlled}. However, despite their strong generalization capabilities, they remain less effective than Seq2Seq and text editing models.

\subsection{Arabic Grammatical Error Correction}
Arabic exhibits a diglossic \cite{Ferguson:1959:diglossia} linguistic nature where a non-standard variety, Dialectal Arabic (DA), coexists with Modern Standard Arabic (MSA), the standard form of the language.

\paragraph{MSA GEC} 
% The first major efforts on Arabic GEC for MSA were initiated by the Qatar Arabic Language Bank (QALB) project \cite{zaghouani-etal-2014-large,zaghouani-etal-2015-correction}, which organized the first Arabic GEC shared tasks: QALB-2014 \cite{mohit-etal-2014-first} and QALB-2015 \cite{rozovskaya-etal-2015-second}. More recently, \newcite{habash-palfreyman-2022-zaebuc} introduced the ZAEBUC corpus, a dataset of essays written by native (L1) Arabic-speaking university students. Just like in English and other languages, multiple approaches have been proposed for MSA GEC including feature-based classifiers \cite{rozovskaya-etal-2014-columbia,farra-etal-2014-generalized,bougares-bouamor-2015-ummu,nawar-2015-cufe} and NMT-based systems \cite{watson-etal-2018-utilizing,solyman-etal-2021-synthetic,solyman-etal-2022-automatic,solyman-etal-2023-optimizing}. More recently, \newcite{kwon-etal-2023-beyond} evaluated the performance of LLMs on Arabic GEC and adapted GECToR \cite{omelianchuk-etal-2020-gector}, an edit-based model originally developed for English, to Arabic GEC, however, they weren't successful in making the edit-based model work for Arabic. The current SOTA in Arabic GEC was established by \newcite{alhafni-etal-2023-advancements}, who benchmarked various pretrained Arabic Seq2Seq models and showed that applying contextualized morphological preprocessing and integrating grammatical error detection (GED) information in Seq2Seq models improves performance. Their approach achieved SOTA results on QALB-2014, QALB-2015, and ZAEBUC.
The first major efforts on MSA GEC were initiated by the Qatar Arabic Language Bank (QALB) project \cite{zaghouani-etal-2014-large,zaghouani-etal-2015-correction}, which organized the QALB-2014 \cite{mohit-etal-2014-first} and QALB-2015 \cite{rozovskaya-etal-2015-second} shared tasks. More recently, \newcite{habash-palfreyman-2022-zaebuc} introduced the ZAEBUC corpus, a dataset of essays written by native Arabic-speaking university students. Approaches to MSA GEC have included feature-based classifiers \cite{rozovskaya-etal-2014-columbia,farra-etal-2014-generalized,bougares-bouamor-2015-ummu,nawar-2015-cufe} and NMT-based systems \cite{watson-etal-2018-utilizing,solyman-etal-2021-synthetic,solyman-etal-2022-automatic,solyman-etal-2023-optimizing}. LLMs have also been evaluated for MSA GEC \cite{kwon-etal-2023-beyond,alhafni-etal-2023-advancements,magdy-etal-2024-gazelle}, but attempts to adapt text editing models have been largely ineffective. The current SOTA was established by \newcite{alhafni-etal-2023-advancements}, who incorporated contextualized morphological preprocessing and grammatical error detection (GED) features into Seq2Seq models, achieving SOTA results on the QALB-2014, QALB-2015, and ZAEBUC datasets.

\paragraph{DA GEC}

Dialectal Arabic (DA) comprises multiple regional varieties that differ from MSA and each other in phonology, morphology, and lexicon. While primarily spoken, DA lacks standardized orthography, though its written use has grown on social media, where it appears in varied and noisy forms. To address this, \newcite{habash-etal-2012-conventional,habash-etal-2018-unified} introduced the Conventional Orthography for Dialectal Arabic (CODA), a standardized spelling convention for DA. CODA has since been used to develop multiple DA datasets \cite{habash-etal-2012-morphological,eskander-etal-2013-processing,ARZTB:2014,diab-etal-2014-tharwa,pasha-etal-2014-madamira,Jarrar:2016:curras,khalifa-etal-2018-morphologically}. Building on this work, \newcite{eryani-etal-2020-spelling} created the MADAR CODA Corpus, which consists of parallel sentences in CODA and their original raw form for five Arabic city dialects. CODAfication--the process of normalizing DA into CODA--has been addressed using feature-based methods \cite{eskander-etal-2013-processing} and morphological disambiguation models \cite{pasha-etal-2014-madamira,zalmout-etal-2018-noise,khalifa-etal-2020-morphological,zalmout-habash-2020-joint,obeid-etal-2022-camelira}. More recently, \newcite{alhafni-etal-2024-exploiting} framed CODAfication as a DA GEC problem, benchmarking pretrained Arabic Seq2Seq models on the MADAR CODA corpus and demonstrating that incorporating dialect identification improves performance.

% Various approaches have been proposed for CODAfication—the normalization of DA into CODA—including feature-based methods \cite{eskander-etal-2013-processing} and morphological disambiguation models \cite{pasha-etal-2014-madamira,zalmout-etal-2018-noise,khalifa-etal-2020-morphological,zalmout-habash-2020-joint,obeid-etal-2022-camelira}. More recently, \newcite{alhafni-etal-2024-exploiting} framed CODAfication as a DA GEC problem, benchmarking pretrained Arabic Seq2Seq models on the MADAR CORA corpus, showing that incorporating dialect identification improves performance.

\textbf{In this work}, we propose a generalizable and efficient text editing approach and evaluate its effectiveness on both MSA and DA GEC. For MSA GEC, we benchmark our models against \newcite{alhafni-etal-2023-advancements} on QALB-2014, QALB-2015, and ZAEBUC. For DA GEC, we build on \newcite{alhafni-etal-2024-exploiting} by framing CODAfication as a DA GEC problem, evaluating our approach on the MADAR CODA corpus and comparing it to their results.

\begin{figure*}[t]
\centering
\includegraphics[width=\textwidth]{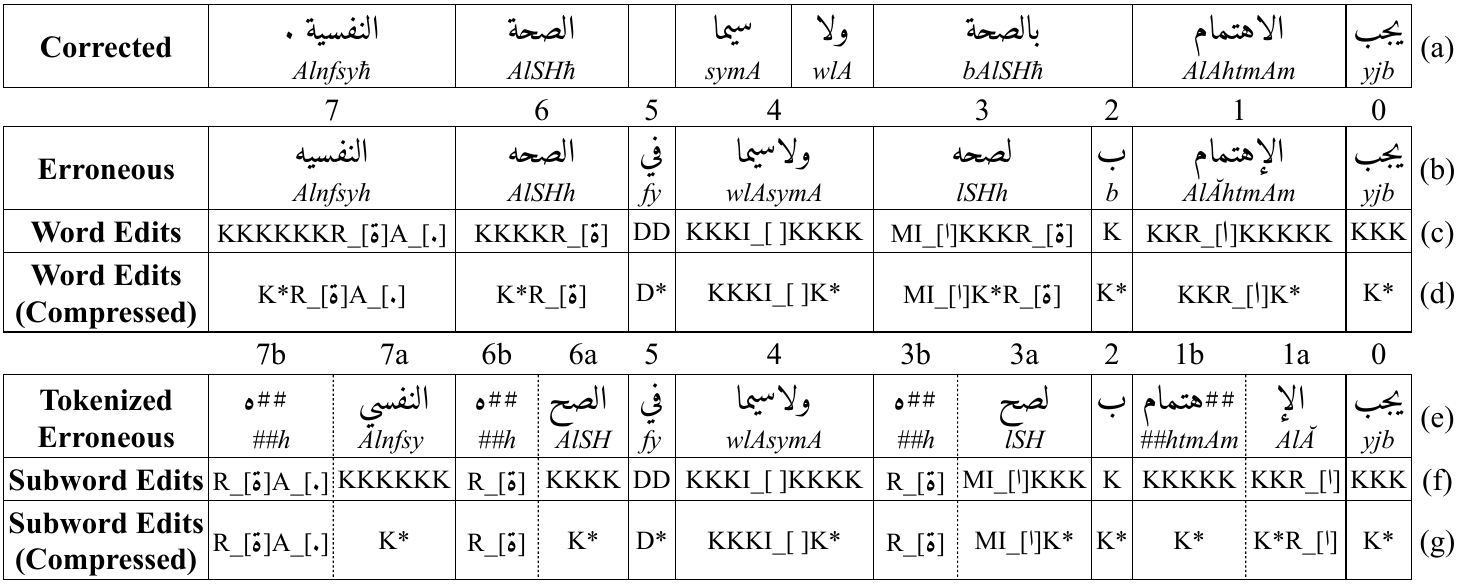}
\caption{
An example showing the different edit representations: words, words (compressed), subwords, and subwords (compressed). The edit operations are keep (\textbf{K}/\textbf{K*}), delete (\textbf{D}/\textbf{D*}), merge before (\textbf{M}), replace (\textbf{R\_[c]}), insert (\textbf{I\_[c]}), and append (\textbf{A\_[c]}). Solid lines indicate word alignments between the corrected and erroneous sentences, while dotted lines denote erroneous subword boundaries. The sentence in the figure can be translated as \textit{``Health, especially mental health, must be taken care of''}.
}
\label{fig:edits}
\end{figure*}

\section{Approach}
We adopt a text editing approach to GEC and frame the task as a sequence tagging problem. Formally, given an input erroneous sequence $x = x_1, x_2, ..., x_n$, the goal is to assign a sequence of edit operations $e = e_1, e_2, ..., e_n$; $e_i \in E$, where $E$ is the edit vocabulary, such that applying edit $e_i$ on the input token $x_i$ at each position $i$ would result in the corrected sequence $y = y_1, y_2, ..., y_m$.
% The choice of edit vocabulary size ($|E|$) presents a key trade-off: a larger vocabulary allows for more precise corrections but increases model complexity, while a smaller one improves learning efficiency at the risk of reduced expressiveness.  In the next two sections, we describe how we extract edits and introduce four methods for controlling $|E|$ while maintaining sufficient coverage.
In the next two sections, we describe how we extract the edits and the edit representations we use to build our edit-based taggers.

\subsection{Edit Extraction}
\label{sec:edit-extraction}
We begin by aligning erroneous and corrected sentence pairs at the word level using a weighted Levenshtein edit distance \cite{Levenshtein:1966:binary}, which represents the minimum number of insertions, deletions, and replacements required to correct the erroneous sentence, with each edit affecting a single word. However, some errors span multiple words. To capture multi-word edits, we follow the approach of \newcite{alhafni-etal-2023-advancements} by extending the alignment process with an iterative algorithm that greedily merges or splits adjacent words, minimizing the overall cumulative edit distance. After obtaining the word-level alignment, we apply the algorithm again, this time to each aligned word pair rather than the entire sentence, to determine character-level alignments. This process identifies the minimal character edits in terms of keep (\textbf{K}), delete (\textbf{D}), merge before (\textbf{M}), insert (\textbf{I\_[c]}), and replace (\textbf{R\_[c]}) that are needed to transform each erroneous word into its correction, where the inserted or replaced character (\textbf{c}) is explicitly specified.

Figure~\ref{fig:edits} presents an example of an aligned erroneous-corrected sentence pair along with the corresponding edits. For instance, in row b, the erroneous word  \<الإهتمام> \textit{Al{\AHAMZADN}htmAm}\footnote{\novocalize
Arabic HSB transliteration \cite{Habash:2007:arabic-transliteration}.} (word 1) requires the edit \texttt{KKR\_[\<ا>]KKKKK} (row c) which consists of eight character edits--one replacement and seven keeps--to produce its corrected form \<الاهتمام> \textit{AlAhtmAm}. Similarly, \<لصحه> \textit{lSHh} (row~b, word~3), must be merged with the word before it, in addition to one insertion and one replacement (\texttt{MI\_[\<ا>]KKKR\_[\<ة>]}, row c).

In some cases, corrections require the insertion of entirely new characters, forming additional words in the erroneous input. Since we frame the task as a sequence tagging problem, we represent these insertions as appends (\textbf{A\_[c]}) to existing edits rather than introducing standalone edits. This ensures that all edits, including word insertions, remain within the tagging framework. For example, to insert a period at the end of the erroneous sentence in Figure~\ref{fig:edits}, we append the tag (\texttt{A\_[.]}) to the edit of the final word (row c, word 7).

\subsection{Edit Representation}
\label{sec:edit-representation}
The edit representation directly influences the size of the edit vocabulary ($|E|$), creating an important trade-off: a larger vocabulary offers more precise corrections but increases model complexity, whereas a smaller vocabulary enhances learning efficiency at the cost of expressiveness. Controlling $|E|$ is crucial to avoid the explosion of possible edits, which is particularly important when working with morphologically rich languages like Arabic. We explore four methods for controlling $|E|$ while maintaining sufficient coverage.

\paragraph{Edit Compression} Once we obtain character-level edits for each word, we compress them into a more compact representation. The motivation behind this transformation is that while different words may undergo the same type of correction, their character-level edits can differ due to variations in word length. For example, in row b of Figure~\ref{fig:edits}, both words 0 and 2 share a keep edit, yet they receive different edit labels because of their length differences (row c). To address this, we introduce a generalized notation for common edit patterns. Consecutive keep (\textbf{K}) and delete (\textbf{D}) operations are represented as \textbf{K*} and \textbf{D*}, respectively. Similarly, consecutive insertions and appends are merged into a single operation, represented as \textbf{I\_[c*]} for insertions and \textbf{A\_[c*]} for appends, indicating the insertion or appending of multiple characters.

Since there are multiple ways to compress an edit sequence, we select the optimal strategy based on the frequency distribution of edit patterns in the training data. This approach ensures that the most common transformations are encoded in a way that balances expressiveness with efficiency, resulting in a more structured and learnable edit representation.

% https://docs.google.com/spreadsheets/d/1LPT-S8FZBSbl85r02KGMECYCsPt9POgQSjuTLpdsdWc/edit?gid=548125526#gid=548125526
\begin{table}[t]
\setlength{\tabcolsep}{3pt}
\small
\centering
\begin{tabular}{lcccccc}
\toprule
\bf Input & \bf Comp.  & \bf Subset  & \bf Prune  & \bf Edits & \bf OOV\%  & \bf F\textsubscript{0.5} \\\hline

Word & \XSolidBrush & All & - & 16,221  & 1.00\% & 98.4 \\
Subword & \XSolidBrush & All & - & 9,060  &  0.36\% & 98.7 \\\hline

Word & \Checkmark & All & - & 10,410 &  1.00\% & 98.4 \\
Subword & \Checkmark & All & - & 6,170  &  0.36\% & 98.7 \\\hline

Subword & \Checkmark & NoPnx & - & 4,799  &  0.27\% & 98.8 \\
Subword & \Checkmark & Pnx & - & 160 & 0.01\% & 99.4\\\hline

Subword & \Checkmark & All & 10 & 683 & 0.75\% & 98.1 \\
Subword & \Checkmark & All & 20 & 442 & 1.02\% & 97.7 \\
Subword & \Checkmark & All & 30 & 329 & 1.24\% & 97.4 \\\hline

Subword & \Checkmark & NoPnx & 10 & 520 & 0.56\% & 98.2 \\
Subword & \Checkmark & NoPnx & 20 & 335 & 0.75\% & 97.8 \\
Subword & \Checkmark & NoPnx & 30 & 250 & 0.92\% & 97.5 \\\hline

Subword & \Checkmark & Pnx & 10 & 48  & 0.02\% &  99.4 \\
Subword & \Checkmark & Pnx & 20 & 35 &  0.05\% & 99.4 \\
Subword & \Checkmark & Pnx & 30 & 29 &  0.05\% & 99.3\\

\bottomrule

\end{tabular}
\caption{Edit statistics on QALB-2014. \textbf{Input} is the input unit (word or subword). \textbf{Comp.} indicates whether the edit is compressed. \textbf{Subset} specifies whether the edits capture all errors, punctuation-only errors (Pnx), or non-punctuation errors (NoPnx). \textbf{Edits} represents the total number of unique edits in the training set. \textbf{OOV\%} is the percentage of out-of-vocabulary edits (non-unique) in the Dev set of QALB-2014.}
\label{tab:edit-coverage}
\end{table}

\paragraph{Input Unit}
Since Transformer-based models operate at the subword level, we project character-level edits onto subwords while maintaining their boundaries to ensure proper alignment. This not only ensures consistency with the model's input representation but also helps reduce the edit vocabulary size. Our approach is inspired by the method of \newcite{straka-etal-2021-character}, but it differs in several key aspects: (1) \newcite{straka-etal-2021-character} tokenize the erroneous and corrected sentence pairs before aligning them to extract the edits at the subword level. In contrast, our method extracts edits at the word level and then projects them onto subwords; (2) They limit the number of character-level edits per subword edit, while our approach imposes no such restrictions, allowing for broader coverage.

Figure~\ref{fig:edits} presents the subword-level edits in both their uncompressed (row f) and compressed (row g) forms. In the uncompressed subword-level edits, we observe that two subwords (3b and 6b in row e), which belong to different words, share the same edit (\texttt{R\_[<ة>]}). In the compressed representation, we notice that several subwords--such as 0, 1b, 2, 6a, and 7a--end up sharing the same edit (\texttt{K*}).

\paragraph{Edit Segregation} Both the MSA GEC datasets we report on, QALB-2014 and ZAEBUC, exhibit high frequencies of punctuation errors, with punctuation accounting for 40\% of the errors in QALB-2014 and 15\% in ZAEBUC training sets \cite{alhafni-etal-2023-advancements}. To reduce the number of edits that the MSA GEC models must learn, we \textit{segregate} punctuation edits from non-punctuation edits. This results in two versions of the data: one where only non-punctuation errors are tagged, and another where all non-punctuation errors are corrected, leaving only punctuation errors for the model to focus on.  Note that this separation is applied only to the MSA GEC datasets we report on, and not to the DA GEC dataset. Additionally, this approach requires training two systems to be applied sequentially during inference: the first system fixes non-punctuation errors, while the second system addresses only punctuation errors.

\paragraph{Edit Pruning} Morphologically rich languages, in particular, tend to have many infrequent edits in GEC datasets. To improve the model's learning ability, we analyze the distribution of edits in the training data and prune those that occur less frequently than a threshold $T$, replacing them with the ``keep'' edit. This pruning is applied exclusively during training, enabling the model to focus on frequent and informative edits. % while preserving all edit types for evaluation.

\subsection{Edit Coverage}

Table~\ref{tab:edit-coverage} presents edit statistics for QALB-2014, illustrating the impact of our strategies to reduce the edit vocabulary size $|E|$ on edit coverage and upper-bound (oracle) performance on the development (Dev) set. Edit coverage measures the proportion of training edits found in the Dev set, while oracle performance is evaluated using the MaxMatch (M\textsuperscript{2}) scorer \cite{dahlmeier-ng-2012-better} F\textsubscript{0.5} (\S\ref{sec:evaluation}). We use AraBERTv02 \cite{antoun-etal-2020-arabert} for subword tokenization, as it yielded the best results among our tested models (more details in \S\ref{sec:results}).

%Going from word-level to subword-level edits reduces the number of unique training edits from 16,221 to 9,060 (44\% reduction), while also lowering the out-of-vocabulary (OOV) rate in the Dev set from 1\% to 0.4\%. This results in a modest F\textsubscript{0.5} improvement of 0.3 points. Edit compression further reduces the number of unique training edits while maintaining the same OOV\% and oracle performance.

Switching from word-level to subword-level edits reduces unique training edits by 44\% (16,221 to~9,060) and lowers the Dev set OOV rate from 1\% to~0.4\%, yielding a~0.3-point F\textsubscript{0.5} gain. Edit compression further reduces unique edits while preserving OOV\% and oracle performance.

% I dont like this, maybe rewrite?
%Segregating punctuation (Pnx) from non-punctuation (NoPnx) edits further lowers the total number of combined training edits (4,799+160 from 6,170). However, the NoPnx results are obtained after explicitly removing punctuation, making them not directly comparable to the other settings. The Pnx F\textsubscript{0.5} scores are also notably higher because they are evaluated on a version of the Dev set where all non-punctuation errors have already been corrected, effectively creating an easier test case.
Segregating punctuation (Pnx) from non-punctuation (NoPnx) edits reduces combined training edits (4,799+160 from 6,170). However, NoPnx results are not directly comparable, as punctuation is explicitly removed before the evaluation. Pnx F\textsubscript{0.5} scores are higher as they are evaluated on a Dev set with non-punctuation errors already corrected, making the test easier.

To assess the impact of pruning, we apply frequency thresholds of 10, 20, and~30 to remove low-frequency edits. As expected, pruning reduces the number of unique training edits and increases the OOV\% in the Dev set, yet F\textsubscript{0.5} remains largely unaffected. This suggests that the majority of the 6,170 compressed subword edits occur infrequently and contribute little to the model's upper-bound performance. A similar trend is observed for both Pnx and NoPnx edits, reinforcing the idea that many low-frequency edits can be pruned without degrading oracle performance.

We present the same analysis on all datasets in Appendix~\ref{sec:appendix-edits-coverage} Table~\ref{tab:edit-coverage-all}.

\section{Experimental Setup}

\subsection{Data}

\paragraph{MSA GEC} We report on three publicly available MSA GEC datasets. The first is the QALB-2014 shared task dataset~\cite{mohit-etal-2014-first}, followed by the native (L1) test set from the QALB-2015 shared task~\cite{rozovskaya-etal-2015-second}. The third dataset is ZAEBUC~\cite{habash-palfreyman-2022-zaebuc}. QALB-2014 and the L1 test set of QALB-2015 contain comments by native speakers from the Aljazeera news website, whereas ZAEBUC consists of essays written by native university students. We use the publicly available splits for QALB-2014 and QALB-2015, while for ZAEBUC, we use the splits created by \newcite{alhafni-etal-2023-advancements}.

\paragraph{DA GEC} We use the MADAR CODA corpus \cite{eryani-etal-2020-spelling}, a set of 10,000 sentences from five Arabic city dialects (Beirut, Cairo, Doha, Rabat, and Tunis) written in the CODA standard in parallel with their original raw form. The sentences come from the Multi-Arabic Dialect Applications and Resources (MADAR) Project \cite{bouamor-etal-2018-madar} and are in parallel across the cities (2,000 sentences per city). We use the publicly available splits created by \newcite{alhafni-etal-2024-exploiting}.

%Table~\ref{tab:data-stats} presents a summary of the dataset statistics. 
Table~\ref{tab:data-stats} summarizes the dataset statistics.

\subsection{Evaluation}
\label{sec:evaluation}
We use the MaxMatch (M\textsuperscript{2}) scorer \cite{dahlmeier-ng-2012-better}, which evaluates GEC systems by comparing hypothesis edits with reference edits, calculating precision (P), recall (R), F\textsubscript{1}, and F\textsubscript{0.5} scores. F\textsubscript{0.5} weighs precision twice as much as recall, to prioritize the accuracy of edits relative to all edits made by the system.

% ADD IN CAMERA READY
% We use the optimized version of the M\textsuperscript{2} scorer presented by \newcite{alhafni-etal-2023-advancements} that deals with the extreme running times of the original release in cases where the generated outputs differ significantly from the input.

% https://docs.google.com/spreadsheets/d/1LPT-S8FZBSbl85r02KGMECYCsPt9POgQSjuTLpdsdWc/edit?gid=1565831290#gid=1565831290

\begin{table}[t]
\setlength{\tabcolsep}{3.5pt}
\small
\centering
\begin{tabular}{l l l l l l}
\toprule
\bf Dataset & \bf Split  & \bf Lines  & \bf Words  &\bf Err.\%  & \bf Domain \\
\midrule
\multirow{3}{*}{\bf QALB-2014} 
    & Train & 19K & 1M   & 30\% & Comments \\
    & Dev   & 1K & 54K   & 31\% & Comments \\
    & Test  & 968 & 51K   & 32\% & Comments \\
\midrule
{\bf QALB-2015} & Test & 920 & 49K & 27\% & Comments \\
\midrule
\multirow{3}{*}{\bf ZAEBUC}  
    & Train & 150 & 25K   & 24\% & Essays \\
    & Dev   & 33 & 5K   & 25\% & Essays \\
    & Test  & 31 & 5K    & 26\% & Essays \\
\midrule
\multirow{3}{1.3cm}{\bf MADAR CODA}
    & Train & 7K & 40K   & 22\% & Comments \\
    & Dev   & 1.5K & 9K    & 20\% & Comments \\
    & Test  & 1.5K & 9K    & 21\% & Comments \\
\bottomrule
\end{tabular}
%\caption{Corpus statistics of MSA (first three rows) and DA (last row) GEC datasets.}
\caption{Corpus statistics of MSA (QALB, ZAEBUC) and DA (MADAR CODA) GEC datasets.}
\label{tab:data-stats}
\end{table}

\begin{table*}[t]
\setlength{\tabcolsep}{4pt}
% \small
\centering
\begin{tabular}{l  cccc | cccc}
\toprule

& \multicolumn{4}{c|}{\bf QALB-2014} & \multicolumn{4}{c}{\bf ZAEBUC} \\
% \midrule
& {\bf P} & {\bf R} & {\bf F\textsubscript{1}} & {\bf F\textsubscript{0.5}}\phantom{\textsuperscript{\textdagger}} & {\bf P} & {\bf R} & {\bf F\textsubscript{1}} & {\bf F\textsubscript{0.5}}\phantom{\textsuperscript{\textdagger}}\\
\hline
A'2023 (Seq2Seq) & 83.2 & 64.9 & 72.9 & 78.7\phantom{\textsuperscript{\textdagger}}  & 87.3 & 70.6 & 78.1 & 83.4\phantom{\textsuperscript{\textdagger}} \\
A'2023 (Seq2Seq++) & 83.1 & 67.9 & 74.7 & 79.6\phantom{\textsuperscript{\textdagger}} & \underline{87.6} & 73.9 & 80.2 & \underline{84.5}\phantom{\textsuperscript{\textdagger}} \\
\hline
GPT-3.5-turbo & 68.6 & 58.6 & 63.2 & 66.3\phantom{\textsuperscript{\textdagger}} & 71.0 & 63.5 & 67.1 & 69.4\phantom{\textsuperscript{\textdagger}}\\
GPT-4o & 80.7 & 65.7 & 72.4 & 77.2\phantom{\textsuperscript{\textdagger}}  & 86.5 & \textbf{\underline{76.8}} & \textbf{\underline{81.3}} & 84.3\phantom{\textsuperscript{\textdagger}}\\
Fanar & 69.7 & 63.7 & 66.6 & 68.4\phantom{\textsuperscript{\textdagger}}  & 76.3 & 73.6 & 74.9 & 75.8\phantom{\textsuperscript{\textdagger}}\\
Jais-13B-Chat & 49.1 & 36.9 & 42.1 & 46.0\phantom{\textsuperscript{\textdagger}}  & 50.2 & 19.7 & 28.3 & 38.4\phantom{\textsuperscript{\textdagger}}\\
\hline
\textsc{sweet} & 81.8 & 68.8 & 74.7 & 78.8\phantom{\textsuperscript{\textdagger}}  & 85.8 & 72.3 & 78.4 & 82.7\phantom{\textsuperscript{\textdagger}}\\
$\textsc{sweet}^{2}$& 81.9 & \textbf{\underline{70.4}} & \underline{75.7} & 79.3\phantom{\textsuperscript{\textdagger}}  & 85.8 & 73.3 & 79.1 & 83.0\phantom{\textsuperscript{\textdagger}}\\
$\textsc{sweet}^{2}_{\text{NoPnx}}$ + $\textsc{sweet}^{}_{\text{Pnx}}$ & \underline{83.7} & 68.8 & 75.6 & \underline{80.3}\textsuperscript{$\dagger$} & 86.7 & 73.9 & 79.8 & 83.8\phantom{\textsuperscript{\textdagger}}\\
\hline\hline
3-Ensemble & 84.9 & 68.8 & \textbf{76.0} & 81.1\phantom{\textsuperscript{\textdagger}}  & 89.6 & 72.8 & 80.3 & 85.6\phantom{\textsuperscript{\textdagger}}\\
4-Ensemble & \textbf{89.1} & 61.6 & 72.8 & \textbf{81.8}\textsuperscript{$\ddagger$} & \textbf{93.3} & 68.3 & 78.9 & \textbf{86.9}\textsuperscript{$\ddagger$}\\
\bottomrule

\end{tabular}
\caption{MSA GEC results on the Dev sets of QALB-2014 and ZAEBUC. A'2023 is \newcite{alhafni-etal-2023-advancements}. Best non-ensemble results are underlined; best overall results are in bold. $\dagger$ denotes statistical significance over the best baseline; $\ddagger$ denotes statistical significance over both the best baseline and the best non-ensemble model.}
\label{tab:gec-dev-res}
\vspace{-5pt}
\end{table*}

\begin{table}[t]
\setlength{\tabcolsep}{4pt}
% \small
\centering
\begin{tabular}{l  cccc}
\toprule

% & \multicolumn{5}{c}{\bf MADAR-CODA}  \\
% \midrule
& {\bf P} & {\bf R} & {\bf F\textsubscript{1}} & {\bf F\textsubscript{0.5}}\phantom{\textsuperscript{\textdagger}}  \\
\hline
A'2024 (Seq2Seq) & 86.8 & 77.4 & 81.8 & 84.7\phantom{\textsuperscript{\textdagger}} \\
A'2024 (Seq2Seq++) & 87.6 & \textbf{\underline{79.3}} & \underline{83.3} & 85.8\phantom{\textsuperscript{\textdagger}}  \\\hline
GPT-3.5-turbo & 35.5 & 29.7 & 32.3 & 34.1\phantom{\textsuperscript{\textdagger}}  \\
GPT-4o & 53.7 & 54.4 & 54.1 & 53.8\phantom{\textsuperscript{\textdagger}}  \\
Fanar & 24.5 & 28.8 & 26.4 & 25.2\phantom{\textsuperscript{\textdagger}}   \\
Jais-13B-Chat & 14.1 & 15.0 & 14.5 & 14.3\phantom{\textsuperscript{\textdagger}}   \\\hline
\textsc{sweet} & \underline{89.1} & 75.5 & 81.7 & \underline{86.0}\phantom{\textsuperscript{\textdagger}}  \\
$\textsc{sweet}^{\text{2}}$ & 87.5 & 73.5 & 79.9 & 84.3\phantom{\textsuperscript{\textdagger}}   \\\hline\hline
3-Ensemble & 91.7 & 77.4 & \textbf{83.9} & 88.4\phantom{\textsuperscript{\textdagger}}  \\
4-Ensemble & \textbf{93.8} & 72.5 & 81.8 & \textbf{88.6}\textsuperscript{$\ddagger$}   \\
\bottomrule

\end{tabular}
\caption{DA GEC results on the MADAR CODA Dev set. A'2024 is \newcite{alhafni-etal-2024-exploiting}. Best non-ensemble results are underlined; best overall results are in bold. $\ddagger$ denotes statistical significance over both the best baseline and the best non-ensemble model.}
\vspace{-5pt}

\label{tab:coda-dev-res}
\end{table}

\subsection{Models}

% \begin{itemize}
%     \item There are many Arabic BERT models available (cite papers)
%     \item We chose to experiment with best three Arabic BERT models on multiple tasks according to \cite{inoue-etal-2021-interplay}: CAMeLBERT-MSA, AraBERTv02, and ARBERTv2.
%     \item We run multiple experiments to measure the effectiveness of the edit representation. We found that AraBERTv02 is the best performer \textcolor{red}{We have to mention pruning}
%     \item We estimate an upper bound of GEC performance to see how we much the train edits cover the Dev edits
%     \item for QALB-2014 and Zaebuc, we have two models: once to fix no-punctuation errors and the other to fix  punctuation errors. 
%     \item for QALB-2014, we use the shared task train data. For Zaebuc, we used QALB-2014 and Zaebuc 10. We over sampled Zaebuc 10 times to account for the domain shift in QALB-2014. This is different from what was done by \cite{alhafni-etal-2023-advancements} where they used the training data of QALB-2014,  QALB-2015 (L2), and Zaebuc. For CODAfication, we use the training data for the MADAR-CODA corpus.
% \end{itemize}

% \textcolor{red}{Explain that we include LLMs but it's not a fair comparison due to cost, data contamenation, etc.}

\paragraph{LLMs} We evaluate four LLMs: two commercial models and two open-source, Arabic-centric models. The commercial models include OpenAI's GPT-3.5-turbo and GPT-4o \cite{gpt4}, while the Arabic-centric models are Jais-13B-Chat \cite{sengupta2023jais} and the recently introduced Fanar LLM \cite{fanar2025}. We prompt GPT-3.5-turbo, GPT-4o, and Fanar through the OpenAI API, while Jais-13B-Chat is prompted using Hugging Face's Transformers \cite{wolf-etal-2020-transformers}. Our experiments use both English and Arabic prompts, employing 0-shot and 5-shot prompting strategies. We design the prompts to elicit minimal edit-style corrections, ensuring that the LLMs' outputs remain as close as possible to the original input in phrasing and lexical choices. We present our prompts in Figures~\ref{fig:zero-shot-prompts} and \ref{fig:few-shot-prompts} in Appendix~\ref{sec:appendix-prompts}.

% while the Arabic-centric models are Jais-13B-Chat and the recently introduced Fanar LLM.
% We evaluate four LLMs: two commercial models and two open-source, Arabic-centric models. The commercial models include OpenAI's GPT-3.5-turbo and GPT-4o \cite{gpt4}, while the Arabic-centric models are instruction-tuned variants, chosen to align with our prompt templates, which include explicit instructions. Specifically, we use Jais-13B-Chat \cite{sengupta2023jais} and the recently introduced Fanar LLM \cite{fanar2025}. 

\paragraph{Edit Taggers} To investigate the impact of edit representation design on performance (\S\ref{sec:edit-representation}), we build edit taggers with different configurations. For word-level tagging, we use the representation of the first subword of each word and pass it through the subsequent layers. For subword-level tagging, we use the representation of each subword individually. Several Arabic pretrained transformer encoders based on BERT \cite{devlin-etal-2019-bert} have been developed \cite{antoun-etal-2020-arabert,abdul-mageed-etal-2021-arbert,inoue-etal-2021-interplay,ghaddar-etal-2022-revisiting}. We select the three best-performing Arabic BERT models, as identified by \newcite {inoue-etal-2021-interplay} across various sentence and token classification tasks: AraBERTv02 \cite{antoun-etal-2020-arabert}, ARBERTv2 \cite{abdul-mageed-etal-2021-arbert}, and CAMeLBERT-MSA~\cite{inoue-etal-2021-interplay}.

% Several Arabic pretrained transformer encoders based on BERT \cite{devlin-etal-2019-bert} have been developed, including AraBERT \cite{antoun-etal-2020-arabert}, ARBERT \cite{abdul-mageed-etal-2021-arbert}, CAMeLBERT \cite{inoue-etal-2021-interplay}, QARIB \cite{abdelali2021pretraining}, and JABER \cite{ghaddar-etal-2022-revisiting}. We select the three best-performing Arabic BERT models, as identified by \newcite {inoue-etal-2021-interplay} across various sentence and token classification tasks: AraBERTv02, ARBERTv2, and CAMeLBERT-MSA.

% When evaluating on QALB-2014, our edit taggers are trained exclusively on QALB-2014, following the restrictions set by the QALB-2014 shared task. Similarly, when evaluating on QALB-2015 (L1), we train only on QALB-2014 to maintain consistency. For ZAEBUC, we train on both QALB-2014 and ZAEBUC but upsample ZAEBUC tenfold to compensate for its smaller size and the domain shift between the two datasets. For DA GEC, we train our models using the training split of the MADAR CODA dataset.

For QALB-2014, our edit taggers are trained exclusively on QALB-2014, following the shared task restrictions. For QALB-2015 (L1), we train only on QALB-2014 for consistency. For ZAEBUC, we train on both QALB-2014 and ZAEBUC, upsampling ZAEBUC tenfold to address its smaller size and domain shift. For DA GEC, we train on the MADAR CODA training split. The hyperparameters we used are detailed in Appendix~\ref{sec:hyperparams}. 

% We use Hugging Face's Transformers to build our edit taggers. The hyperparameters we used are detailed in Appendix~\ref{sec:hyperparams}. 

\begin{table*}[t]
\setlength{\tabcolsep}{3.8pt}
% \small
\centering
\begin{tabular}{l  cccc | cccc | cccc}
\toprule

& \multicolumn{4}{c|}{\bf QALB-2014} & \multicolumn{4}{c|}{\bf QALB-2015} & \multicolumn{4}{c}{\bf ZAEBUC} \\
% \midrule
& {\bf P} & {\bf R} & {\bf F\textsubscript{1}} & {\bf F\textsubscript{0.5}}\phantom{\textsuperscript{\textdagger}} & {\bf P} & {\bf R} & {\bf F\textsubscript{1}} & {\bf F\textsubscript{0.5}}\phantom{\textsuperscript{\textdagger}} & {\bf P} & {\bf R} & {\bf F\textsubscript{1}} & {\bf F\textsubscript{0.5}}\phantom{\textsuperscript{\textdagger}} \\
\hline
A'2023 (Seq2Seq) & 84.0 & 64.7 & 73.1  & 79.3\phantom{\textsuperscript{\textdagger}} & 82.0 & 71.7 & 76.5 & 79.7\phantom{\textsuperscript{\textdagger}} & \underline{86.0} & 71.6 & 78.2 & 82.7\phantom{\textsuperscript{\textdagger}} \\
A'2023 (Seq2Seq++) & 84.2 & 65.4 & 73.6 & 79.6\phantom{\textsuperscript{\textdagger}} & \underline{82.6} & 72.1 & 77.0 & \underline{80.3}\phantom{\textsuperscript{\textdagger}} & 85.9 & 73.4 & 79.2 & \underline{83.1}\phantom{\textsuperscript{\textdagger}} \\
\hline
GPT-4o & 81.5 & 65.5 & 72.6 & 77.7\phantom{\textsuperscript{\textdagger}} & 81.1 & \textbf{\underline{74.3}} & 77.5 & 79.6\phantom{\textsuperscript{\textdagger}} & 84.4 & \textbf{\underline{75.9}} & \underline{79.9} & 82.5\phantom{\textsuperscript{\textdagger}} \\
\hline
$\textsc{sweet}^{2}$& 82.6 & \textbf{\underline{69.5}} & \textbf{\underline{75.5}} & 79.6\phantom{\textsuperscript{\textdagger}} & 80.0 & \textbf{\underline{74.3}} & 77.0 & 78.8\phantom{\textsuperscript{\textdagger}} & 85.5 & 74.4 & 79.6 & 83.0\phantom{\textsuperscript{\textdagger}} \\
$\textsc{sweet}^{2}_{\text{NoPnx}}$ + $\textsc{sweet}^{}_{\text{Pnx}}$ & \underline{84.5} & 67.7 & 75.2 & \underline{80.5}\textsuperscript{$\dagger$} & 82.2 & 73.6 & \underline{77.7} & \underline{80.3}\phantom{\textsuperscript{\textdagger}} & 85.7 & 74.1 & 79.5 & \underline{83.1}\phantom{\textsuperscript{\textdagger}} \\
\hline\hline
3-Ensemble & 85.7 & 67.4 & 75.4 & 81.3\phantom{\textsuperscript{\textdagger}} & 83.7 & 73.3 & \textbf{78.1} & 81.3\phantom{\textsuperscript{\textdagger}} & 89.7 & 73.7 & \textbf{80.9} & 85.9\phantom{\textsuperscript{\textdagger}} \\
4-Ensemble& \textbf{89.7} & 60.2 & 72.0 & \textbf{81.7}\textsuperscript{$\ddagger$} & \textbf{88.3} & 66.7 & 76.0 & \textbf{82.9}\textsuperscript{$\ddagger$} & \textbf{93.4} & 68.9 & 79.3 & \textbf{87.2}\textsuperscript{$\ddagger$} \\
\bottomrule

\end{tabular}
\caption{MSA GEC results on the Test sets of QALB-2014, QALB-2015 (L1), and ZAEBUC. A'2023 is \newcite{alhafni-etal-2023-advancements}. Best non-ensemble results are underlined; best overall results are in bold. $\dagger$ denotes statistical significance over the best baseline; $\ddagger$ denotes statistical significance over both the best baseline and the best non-ensemble model.}
\label{tab:gec-test-res}
\end{table*}

% https://docs.google.com/spreadsheets/d/1LPT-S8FZBSbl85r02KGMECYCsPt9POgQSjuTLpdsdWc/edit?gid=1373417432#gid=1373417432
% \begin{table}[t]
% \setlength{\tabcolsep}{3.5pt}
% % \small
% \centering
% \begin{tabular}{l  lllll}
% \toprule

% % & \multicolumn{5}{c}{\bf MADAR-CODA}  \\
% % \midrule
% & {\bf P} & {\bf R} & {\bf F\textsubscript{1}} & {\bf F\textsubscript{0.5}} & {\bf WER}  \\
% \hline
% A'2024 (S2S) & 86.8 & 77.4 & 81.8 & 84.7 & 0.062 \\
% A'2024 (S2S++) & 87.6 & 79.3 & 83.3 & 85.8 & 0.057 \\\hline
% GPT-4o & 53.7 & 54.4 & 54.1 & 53.8 & 0.189 \\\hline
% \textsc{sweet} & 87.5 & 73.5 & 79.9 & 84.3 & 0.071  \\\hline\hline
% 3-Ensemble & 91.7 & 77.4 & 83.9 & 88.4 & 0.060  \\
% + GPT-4o & 93.8 & 72.5 & 81.8 & 88.6 & 0.071  \\
% \hline

% \end{tabular}
% \caption{CODAfication results on the Test set of MADAR-CODA. A'2024 refers to \newcite{alhafni-etal-2024-exploiting}. \textcolor{red}{The ensemble here includes the best CAMeLBERT model. Should we report test results on CAMeLBERT separately?}}
% \label{tab:data-stats}
% \end{table}

\begin{table}[t]
\setlength{\tabcolsep}{4pt}
% \small
\centering
\begin{tabular}{l  cccc}
\toprule

% & \multicolumn{5}{c}{\bf MADAR-CODA}  \\
% \midrule
& {\bf P} & {\bf R} & {\bf F\textsubscript{1}} & {\bf F\textsubscript{0.5}}\phantom{\textsuperscript{\textdagger}}   \\
\hline
A'2024 (Seq2Seq) & 87.3 & 78.0 & 82.4 & 85.2\phantom{\textsuperscript{\textdagger}}  \\
A'2024 (Seq2Seq++) & 88.4 & \textbf{\underline{79.0}} & \underline{83.4} & 86.3\phantom{\textsuperscript{\textdagger}}  \\\hline
GPT-4o & 56.1 & 54.8 & 55.5 & 55.9\phantom{\textsuperscript{\textdagger}} \\\hline
\textsc{sweet} & \underline{89.4} & 76.6 & 82.5 & \underline{86.5}\phantom{\textsuperscript{\textdagger}}   \\\hline\hline
3-Ensemble & 92.2 & 77.7 & \textbf{84.3} & \textbf{88.9}\textsuperscript{$\ddagger$}  \\
4-Ensemble & \textbf{94.0} & 72.9 & 82.1 & 88.8\phantom{\textsuperscript{\textdagger}}  \\
\bottomrule

% 		
%	%	%	%
% %	%	%	%

% 89.1%	76.5%	82.3%	86.2%

\end{tabular}
\caption{DA GEC results on the Test set of MADAR CODA. A'2024 is \newcite{alhafni-etal-2024-exploiting}. Best non-ensemble results are underlined, best overall results are in bold. $\ddagger$ denotes statistical significance over both the best baseline and the best non-ensemble model.}
% \textcolor{red}{The ensemble here includes the best CAMeLBERT model. Should we report test results on CAMeLBERT separately?}}
\label{tab:coda-test-res}
\end{table}

\subsection{Ensembling} We construct majority vote ensemble models by aggregating the outputs of multiple GEC systems. This is enabled by our edit extraction algorithm (\S\ref{sec:edit-extraction}), which allows us to align and extract edits from models with different architectures. Using this algorithm, we first align each model's output with the input text, extract the proposed edits, and then determine the final edit sequence through majority voting. Following \citet{tarnavskyi-etal-2022-ensembling}, we retain an edit only if at least $k - 1$ models out of $k$ models predict it; otherwise, we leave the input unchanged. This strategy prioritizes precision over recall, which is crucial for GEC systems, as precision is generally more important than correcting every possible error \cite{bryant-etal-2023-grammatical}.

\section{Results}
\label{sec:results}
%Tables \ref{tab:gec-dev-res} and \ref{tab:coda-dev-res} present the Dev results for MSA and DA GEC, respectively. For MSA GEC, we compare our results with those reported by \newcite{alhafni-etal-2023-advancements}, including their best Seq2Seq models and their best setup, which incorporates morphological preprocessing and GED information (Seq2Seq++). For DA GEC, we compare our results with those from \newcite{alhafni-etal-2024-exploiting}, which also include the best Seq2Seq and their best setup, which uses dialect identification information (Seq2Seq++).

% Tables \ref{tab:gec-dev-res} and \ref{tab:coda-dev-res} show the Dev results for MSA and DA GEC, respectively. For MSA GEC, we compare our results with those of \newcite{alhafni-etal-2023-advancements}, including their best Seq2Seq models and their Seq2Seq++ setup, which incorporates morphological preprocessing and GED information. For DA GEC, we compare with \newcite{alhafni-etal-2024-exploiting}, including their best Seq2Seq model and Seq2Seq++ setup, which integrates dialect identification.

Tables~\ref{tab:gec-dev-res} and~\ref{tab:coda-dev-res} show the Dev results for MSA and DA GEC, respectively. For each dataset, we compare our models with the best-performing Seq2Seq and Seq2Seq++ baselines reported by \newcite{alhafni-etal-2023-advancements} and \newcite{alhafni-etal-2024-exploiting}. The Seq2Seq++ setups incorporate additional signals, such as morphological preprocessing and GED information for MSA GEC, or dialect identification for DA GEC. Full results for all Seq2Seq-based baseline variants across datasets are provided in Appendices~\ref{sec:appendix-gec-res-full-baselines} and~\ref{sec:appendix-coda-res-full-baselines}.

\paragraph{LLMs}
We present LLMs results on MSA and DA GEC using their best setups, optimized for average F\textsubscript{0.5} across all datasets based on prompt language and strategy (0-shot vs. 5-shot). Full results are in Table~\ref{tab:llm-results} (Appendix~\ref{sec:appendix-llms-res}).

For QALB-2014, GPT-4o and Fanar outperform GPT-3.5 and Jais-13B-Chat, with GPT-4o achieving the best performance, though none surpass \newcite{alhafni-etal-2023-advancements}. On ZAEBUC, GPT-4o leads, achieving the highest recall (76.8) and F\textsubscript{1} (81.3). For DA GEC, GPT-4o is the top LLM, but overall LLM performance is notably lower than for MSA.

% https://docs.google.com/spreadsheets/u/1/d/1LPT-S8FZBSbl85r02KGMECYCsPt9POgQSjuTLpdsdWc/edit?gid=44693648#gid=44693648

\setlength\dashlinedash{1pt}
\setlength\dashlinegap{1.5pt}
\setlength\arrayrulewidth{0.3pt}

\begin{table*}[t]
\centering

\setlength{\tabcolsep}{2.5pt}
\begin{tabular}{lccc|ccc|ccc}
    \toprule
        \multicolumn{1}{l}{} & \multicolumn{3}{c}{\textbf{QALB-2014}} & \multicolumn{3}{c}{\textbf{ZAEBUC}} & \multicolumn{3}{c}{\textbf{MADAR CODA}} \\
        \multicolumn{1}{l}{} & \textbf{Baseline} & \textbf{\textsc{sweet}} & \textbf{Ensemble} & \textbf{Baseline} & \textbf{\textsc{sweet}} & \textbf{Ensemble} & \textbf{Baseline} &\textbf{\textsc{sweet}} &  \textbf{Ensemble} \\\hline

        Delete  & 41.1 & \underline{44.3} & \textbf{45.2} & 51.9 & \underline{\textbf{63.6}} & 62.5 & 0.0  & 0.0 & 0.0 \\
        Merge-B & \textbf{\underline{94.0}} & 93.7 & 93.8 & 96.7 & \underline{\textbf{96.9}} & 96.6 & \textbf{\underline{94.4}} & 86.6 & 92.7 \\
        Merge-I & \textbf{\underline{93.8}} & 93.5 & 93.6 & 96.7 & \underline{\textbf{96.9}} & 96.6 & \textbf{\underline{93.6}} & 84.8 & 91.6 \\
        M & \textbf{\underline{33.9}} & 33.6 & 28.6 & 48.6 & \underline{\textbf{50.0}} & 41.7 & \textbf{\underline{82.5}} & 78.0 & 82.4 \\
        M+O &  58.0 & \textbf{\underline{61.0}} & 60.6 & 55.6 & \underline{\textbf{100.0}} & 0.0 & 0.0 & 0.0 & 0.0 \\
        O & 94.3 & \textbf{\underline{94.5}} & 94.4 & \underline{\textbf{94.4}} & \underline{\textbf{94.4}} & 94.1 & \textbf{\underline{92.1}} & 90.2 & 91.4 \\
        O+X & 78.1 & \underline{81.5} & \textbf{83.3} & 0.0 &  0.0 &  0.0 & 0.0 & 0.0 & 0.0 \\
        P & 75.0 & \underline{75.6} & \textbf{76.8} & 62.8 & \underline{\textbf{71.9}} & 70.4 & \textbf{\underline{65.8}} & 35.7 & 55.6 \\
        S & 46.4 & \underline{57.1} & \textbf{57.4} & 40.4 & \underline{\textbf{47.6}} & 46.9 & \textbf{\underline{83.1}} & 82.3 & 83.0 \\
        X & 61.2 & \underline{61.4} & \textbf{62.4} & 72.9 & \underline{\textbf{74.1}} & \textbf{74.1} & 73.8 & \underline{76.9} & \textbf{79.5} \\
        Split & \underline{87.1} & 83.8 & \textbf{87.6} & 88.2 & \underline{90.0} & \textbf{95.2} & \underline{85.9} & 83.3 & \textbf{86.8} \\
        UNK & \underline{\textbf{59.2}} & 55.0 & 56.0 & \underline{\textbf{63.1}} & 44.6 & 47.6 & 93.0 & \textbf{\underline{94.6}} & 94.1 \\
        C & \underline{\textbf{97.1}} & 96.4 & 94.7 & \underline{\textbf{96.1}} & 96.0 & 93.9 & \textbf{\underline{97.0}} & 96.0 & 94.7 \\\hline
        Macro Avg. & 70.7 & \underline{71.6} & \textbf{71.9} & 66.7 & \underline{\textbf{71.2}} & 63.1 & \textbf{\underline{78.3}} & 73.5 & 77.4 \\
    
    \bottomrule
\end{tabular}

\caption{Error type performance on the Dev sets of QALB-2014, ZAEBUC, and MADAR CODA for the best Seq2Seq++ baseline, the best \textsc{sweet} model ($\textsc{sweet}^{2}_{\text{NoPnx}}$ + $\textsc{sweet}^{}_{\text{Pnx}}$ for QALB-2014 and ZAEBUC; $\textsc{sweet}$ for MADAR CODA), and the best ensemble (4-Ensemble). Results are reported in terms of F\textsubscript{0.5}. Best non-ensemble results are underlined; best overall results are in bold. UNK refers to unknown error types; C refers to correct words.}
\label{tab:error-analysis}
\end{table*}

% Specific error type performance of AraBART and our best system (AraBART+Morph+GED13) on average
% on the Dev sets of QALB-2014, QALB-2015, and ZAEBUC. Results are reported in terms of F0.5. The best results are in bold.

\paragraph{Edit Taggers}
%The full set of edit tagging results on the Dev sets, exploring different edit design choices with the CAMeLBERT-MSA, AraBERTv02, and ARBERTv2 models, is provided in Table~\ref{tab:all-dev-results} in Appendix~\ref{sec:appendix-bert-res}. AraBERTv02 consistently outperforms the other models. We find that compressing the edits and representing them at the subword level always improve performance. Pruning low-frequency edits further enhances results, with a threshold of 10 performing best on QALB-2014 and MADAR CODA, and a threshold of 30 performing best on ZAEBUC. 
Table~\ref{tab:all-dev-results} (Appendix~\ref{sec:appendix-bert-res}) presents the full edit tagging results on the Dev sets, exploring edit design choices using CAMeLBERT-MSA, AraBERTv02, and ARBERTv2. AraBERTv02 consistently performs best. Subword-level edits, compression, and pruning improve performance, with optimal pruning thresholds of 10 for QALB-2014 and MADAR CODA, and 30 for ZAEBUC.  

%The optimal setup for each dataset (subword, compression, pruning) is presented in Tables \ref{tab:gec-dev-res} and \ref{tab:coda-dev-res}. We henceforth refer this system as \textsc{sweet} (Subword Edit Error Tagger). \textsc{sweet} achieves an F\textsubscript{0.5} score of 78.8 on QALB-2014 and 86.0 on MADAR CODA, surpassing the Seq2Seq baseline on QALB-2014 and setting a new SOTA on MADAR CODA. On ZAEBUC, \textsc{sweet} reaches an F\textsubscript{0.5} score of 82.7, trailing behind the Seq2Seq baseline.

The optimal setup for each dataset (subword, compression, pruning) is presented in Tables \ref{tab:gec-dev-res} and \ref{tab:coda-dev-res}. We henceforth refer to this system as \textsc{sweet} (Subword Edit Error Tagger). 
\textsc{sweet} achieves an F\textsubscript{0.5} of 78.8 on QALB-2014 and 86.0 on MADAR CODA, outperforming the Seq2Seq baseline on QALB-2014 and setting a new SOTA on MADAR CODA (though the improvement is not statistically significant).\footnote{Statistical significance was done using a two-sided approximate randomization test.} On ZAEBUC, it scores 82.7 F\textsubscript{0.5},  trailing behind the Seq2Seq baseline.

Consistent with previous work on text editing \cite{omelianchuk-etal-2020-gector,straka-etal-2021-character}, we find that iterative correction improves MSA GEC up to two iterations ($\textsc{sweet}^{2}$), achieving 79.3 on QALB-2014 and 83.0 F\textsubscript{0.5} on ZAEBUC, with the highest recall on QALB-2014 (70.4). However, iterative correction degrades DA GEC performance.

Separating non-punctuation edits ($\textsc{sweet}_{\text{NoPnx}}$) from punctuation edits ($\textsc{sweet}_{\text{Pnx}}$) improves MSA GEC performance. The best setup applies these systems in sequence: two iterations of non-punctuation correction followed by one iteration of punctuation correction ($\textsc{sweet}^{2}_{\text{NoPnx}}$ + $\textsc{sweet}^{}_{\text{Pnx}}$). This setup achieves the highest F\textsubscript{0.5} score among text editing models, setting a new SOTA on QALB-2014 with 80.3. This improvement is statistically significant compared to Seq2Seq++ ($p < 0.05$) and is driven by a precision of 83.7. Its performance on ZAEBUC leads other edit tagging techniques but trails behind GPT-4o. 
%(with gains in both recall at 86.7 and precision at 73.9).

%For our ensemble models (3-Ensemble), we combine the outputs of the best three non-LLM models for each dataset. For QALB-2014 and ZAEBUC, we use Seq2Seq++, $\textsc{sweet}^{2}$, and the cascaded setup $\textsc{sweet}^{2}_{\text{NoPnx}}$ + $\textsc{sweet}^{}_{\text{Pnx}}$. For MADAR CODA, we combine Seq2Seq++, \textsc{sweet}, and the second-best $\textsc{sweet}$ model which uses CAMeLBERT-MSA (see Table~\ref{tab:all-dev-results} in Appendix~\ref{sec:appendix-bert-res}). The 3-Ensembles outperform single models, achieving SOTA results across all datasets. The improvements mainly stem from increased precision, with a trade-off in recall. Adding the outputs of GPT-4o to the ensembles further boosts performance, achieving 81.8 F\textsubscript{0.5} on QALB-2014, 86.9 on ZAEBUC, and 88.6 on MADAR CODA.

For our ensemble models (3-Ensemble), we combine the outputs of the top three non-LLM models per dataset. 
For QALB-2014 and ZAEBUC, this includes Seq2Seq++, $\textsc{sweet}^{2}$, and the cascaded setup $\textsc{sweet}^{2}_{\text{NoPnx}}$ + $\textsc{sweet}^{}_{\text{Pnx}}$. For MADAR CODA, we ensemble Seq2Seq++, \textsc{sweet}, and the second-best \textsc{sweet} model using CAMeLBERT-MSA (see Table~\ref{tab:all-dev-results} in Appendix~\ref{sec:appendix-bert-res}).  
The 3-Ensembles outperform single models, achieving SOTA results across all datasets, primarily through increased precision at the cost of recall. Adding GPT-4o's output to the ensemble further boosts performance (4-Ensemble), reaching an F\textsubscript{0.5} of 81.8 on QALB-2014, 86.9 on ZAEBUC, and 88.6 on MADAR CODA. These gains are statistically significant ($p < 0.05$) compared to the best baseline and the best non-ensemble model for each dataset.

\paragraph{Test Results} Tables \ref{tab:gec-test-res} and \ref{tab:coda-test-res} present the Test results for MSA and DA GEC, using the best setups identified from the Dev sets. On QALB-2014, the cascaded setup $\textsc{sweet}^{2}_{\text{NoPnx}}$ + $\textsc{sweet}^{}_{\text{Pnx}}$ sets a new SOTA with 80.5 F\textsubscript{0.5}, outperforming the Seq2Seq and Seq2Seq++ baselines (statistically significant at $p<0.05$). On QALB-2015, this setup matches Seq2Seq++ with an F\textsubscript{0.5} of 80.3. Similarly, on ZAEBUC, it achieves 83.1, on par with Seq2Seq++. On MADAR CODA, $\textsc{sweet}$ achieves 86.5, outperforming Seq2Seq++ (though not statistically significant). Our ensemble models further enhance performance across all datasets, reaching 81.7 on QALB-2014, 82.9 on QALB-2015, 87.2 on ZAEBUC, and 88.9 on MADAR CODA. Notably, adding GPT-4o's output to the ensemble (i.e., the 4-Ensemble) yields statistically significant improvements for MSA GEC. On MADAR CODA, the 3-Ensemble already achieves statistically significant gains, while the addition of GPT-4o does not lead to further improvement.

\subsection{Error Analysis}
Table~\ref{tab:error-analysis} presents specific error type performance over the Dev sets of QALB-2014, ZAEBUC, and MADAR CODA. We conduct automatic error analysis using ARETA \cite{belkebir-habash-2021-automatic}, an automatic MSA error type annotation tool. ARETA defines error types based on seven classes covering: orthography (\textbf{O}), morphology (\textbf{M}), syntax (\textbf{X}), semantics (\textbf{S}), punctuation (\textbf{P}), merges (\textbf{Merge-Beginning}/\textbf{Merge-Inside}), and splits (\textbf{Split}). 
% Table~\ref{tab:error-type-stats} in Appendix~\ref{sec:appendix-error-type-stats} presents the error type statistics for QALB-2014, ZAEBUC, and MADAR CODA. 

On QALB-2014, the cascaded setup $\textsc{sweet}^{2}_{\text{NoPnx}}$ + $\textsc{sweet}^{}_{\text{Pnx}}$ outperforms the Seq2Seq++ baseline on most error types, achieving a macro F\textsubscript{0.5} of 71.6. The 4-Ensemble provides a modest improvement, reaching 71.9. On ZAEBUC, while the cascaded $\textsc{sweet}$ setup does not surpass Seq2Seq++ in overall performance (Table~\ref{tab:gec-dev-res}), it achieves higher scores on most individual error types, with a macro F\textsubscript{0.5} of 71.2. This stems from the skewed distribution of error types in the ZAEBUC Dev set, which is dominated by correct words (\textbf{C}) and frequent errors like \textbf{O} and \textbf{Merge}, categories where both models perform similarly. The error types where the cascaded $\textsc{sweet}$ model excels are relatively infrequent (see Table~\ref{tab:error-type-stats} in Appendix~\ref{sec:appendix-error-type-stats}). Notably, while the 4-Ensemble yields the best overall GEC results, it falls short of Seq2Seq++ and the cascaded $\textsc{sweet}$ model in error-type performance, likely due to prioritizing precision over recall (Table~\ref{tab:gec-dev-res}).

On MADAR CODA, neither \textsc{sweet} nor the 4-Ensemble surpasses Seq2Seq++ in error-type performance. While \textsc{sweet} performs slightly better in terms of DA GEC, the improvement is not statistically significant and is driven primarily by precision; in contrast, Seq2Seq++ claims the highest recall (Table~\ref{tab:coda-dev-res}). The 4-Ensemble further increases precision but at the cost of recall. The high proportion of the unknown (\textbf{UNK}) errors also highlights ARETA's limitations in capturing dialect-specific errors, as it was primarily designed for MSA.

% On MADAR CODA, neither \textsc{sweet} nor the 4-Ensemble outperforms Seq2Seq++ in error-type performance. \textsc{sweet} yields slightly better overall GEC scores, but the gains are not statistically significant and come mainly from precision, while Seq2Seq++ achieves the highest recall (Table~\ref{tab:gec-dev-res}). The 4-Ensemble pushes precision further at the expense of recall. The high rate of \textbf{UNK} errors also reflects ARETA’s limitations in handling dialect-specific errors, as it was designed for MSA.

\begin{table}[t]
\setlength{\tabcolsep}{3pt}
% \small
\centering
\begin{tabular}{lc|c|c}
\toprule

\multicolumn{2}{r|}{\multirow{2}{*}{\textbf{Params}}} & \multicolumn{2}{c}{\textbf{Time}}\\
& & \textbf{Init.} & \textbf{Run} \\
\hline
A’2023 (Seq2Seq) & 139M & 1.7 & 70.7 \\
A’2023 (Seq2Seq++) & 502M & 24.7 & 218.5 \\
\hline
\textsc{sweet} & 135M & 1.3 & 11.6 \\
$\textsc{sweet}^{\text{2}}$ & 135M & 1.3 & 23.2 \\
$\textsc{sweet}^{2}_{\text{NoPnx}}$ + $\textsc{sweet}^{}_{\text{Pnx}}$ & 270M & 2.7 & 34.8 \\
3-Ensemble & 908M & 28.7 & 276.4 \\
\bottomrule

\end{tabular}
\caption{Number of parameters (Params.), initialization time (Init.), and runtime for different models on the Dev set of QALB-2014. Init. and runtime are in seconds and averaged over 10 runs on a single A100 GPU using a batch size of 32.}
 \vspace{-5pt}
\label{tab:runtime}
\end{table}

\subsection{Runtime Performance}
% \begin{itemize}
%     \item We compare our best models to the seq2seq models developed by \cite{alhafni-etal-2023-advancements} in terms of number of parameters, loading time, and inference speed.
%     \item loading time and inference speed where measured over 10 runs using a batch size of 32 on the qalb-2014 dev set on a single A100 GPU.
% \end{itemize}

Table~\ref{tab:runtime} compares our text editing models to the Seq2Seq models from \newcite{alhafni-etal-2023-advancements} in terms of model size, initialization time, and inference runtime. Initialization and inference times were averaged over 10 runs on the QALB-2014 Dev set using a single A100 GPU with a batch size of 32. The reported values for Seq2Seq++ reflect the combined size, initialization, and inference times of all its components. 
Our \textsc{sweet} model is 4x smaller than Seq2Seq++, while the cascaded system $\textsc{sweet}^{2}_{\text{NoPnx}}$ + $\textsc{sweet}^{}_{\text{Pnx}}$ is about half the Seq2Seq++ model size.
In terms of speed, \textsc{sweet} initializes 19x faster than Seq2Seq++, while the cascaded system achieves a 9x initialization speedup. For inference, \textsc{sweet} is also 19x faster, $\textsc{sweet}^{2}$ is 9x faster, and the cascaded setup is 6x faster. Compared to the vanilla Seq2Seq model, \textsc{sweet} runs 6x faster, $\textsc{sweet}^{2}$ is 3x faster, and the cascaded system is twice as fast. Although the 3-Ensemble setup achieves the best performance, it is the largest in model size and the slowest overall.

\section{Conclusion and Future Work}
% \begin{itemize}
%     \item Applying our approach to other languages and dialectal varities (other CODA datasets)
%     \item Use it to generate synthetic data and see how it compares to seq2seq and LLMs
% \end{itemize}

We introduced a data-driven text editing approach that eliminates the need for predefined language-specific edits. By applying it to Arabic, a diglossic and morphologically rich language, we studied the impact of different edit representations on model performance. Our models set new SOTA results on two Arabic GEC benchmarks and matched top-performing systems on two others. Moreover, they offer a significant efficiency advantage, running over six times faster than existing Arabic GEC systems, making them more suitable for practical deployment. We also explored how ensemble models contribute to further performance improvements.

In future work, we plan to extend this approach to other languages and dialectal varieties \cite{Jarrar:2016:curras,khalifa-etal-2018-morphologically} and investigate its potential for generating synthetic data for GEC \cite{Zuchao:2022,zhang-etal-2022-syngec,stahlberg-kumar-2024-synthetic}. We also plan to explore other ensembling approaches \cite{qorib-ng-2023-system,qorib-etal-2022-frustratingly}.

% We introduced a text editing approach that derives edit tags directly from data, removing the need for language-specific edits. We demonstrated its effectiveness on Arabic, a diglossic and morphologically rich language, and examined the impact of different edit representations on model performance. Our models achieved SOTA results on two Arabic GEC benchmarks and performed on par with the best systems on two others. Additionally, they are over six times faster than existing Arabic GEC systems, making them more practical for real-world applications. We also explored the potential of ensemble models to further enhance performance.

% For future work, we aim to extend our approach to other languages and dialectal varieties and explore its use in generating synthetic data for GEC.

\section*{Limitations}
% not evaluating a wide range of LLMs and multilingual transformer encoders
% the context length of BERT
% not doing L2 GEC

While our work demonstrates promising results, there are several considerations that could impact its broader applicability. One limitation is the use of closed-source commercial LLMs, which introduces a degree of uncertainty, as these models may undergo undisclosed updates over time. Such changes could affect the reproducibility of our results. Additionally, we did not report on L2 Arabic GEC, which could provide valuable insights into how our approach generalizes to second-language learners errors.  We also did not explore multilingual transformer encoders, as we hypothesize that monolingual models would be more effective for Arabic GEC. However, future work is needed to verify this assumption. Finally, our analysis focused on Arabic, which may limit the generalizability of our findings to languages with different error correction challenges.

\section*{Ethical Considerations}
 GEC systems can aid in identifying and correcting errors, but they also raise ethical concerns. Misidentifications or miscorrections may frustrate learners, and GEC tools should complement, not replace, human judgment. There is also the risk of malicious use, such as profiling learners based on error patterns, which could lead to bias or privacy issues. It is important to use these systems responsibly to protect end users.

\section*{Acknowledgments}
We thank Ted Briscoe for helpful discussions and constructive feedback. We also acknowledge the support of the High Performance Computing Center at New York University Abu Dhabi.

\bibliography{anthology,camel-bib-v3,custom}

\begin{thebibliography}{90}
\providecommand{\natexlab}[1]{#1}

\bibitem[{Abdul-Mageed et~al.(2021)Abdul-Mageed, Elmadany, and Nagoudi}]{abdul-mageed-etal-2021-arbert}
Muhammad Abdul-Mageed, AbdelRahim Elmadany, and El~Moatez~Billah Nagoudi. 2021.
\newblock \href {https://doi.org/10.18653/v1/2021.acl-long.551} {{ARBERT} {\&} {MARBERT}: Deep bidirectional transformers for {A}rabic}.
\newblock In \emph{Proceedings of the 59th Annual Meeting of the Association for Computational Linguistics and the 11th International Joint Conference on Natural Language Processing (Volume 1: Long Papers)}, pages 7088--7105, Online. Association for Computational Linguistics.

\bibitem[{Alhafni et~al.(2024)Alhafni, Al-Towaity, Fawzy, Nassar, Eryani, Bouamor, and Habash}]{alhafni-etal-2024-exploiting}
Bashar Alhafni, Sarah Al-Towaity, Ziyad Fawzy, Fatema Nassar, Fadhl Eryani, Houda Bouamor, and Nizar Habash. 2024.
\newblock \href {https://doi.org/10.18653/v1/2024.arabicnlp-1.4} {Exploiting dialect identification in automatic dialectal text normalization}.
\newblock In \emph{Proceedings of The Second Arabic Natural Language Processing Conference}, pages 42--54, Bangkok, Thailand. Association for Computational Linguistics.

\bibitem[{Alhafni et~al.(2023)Alhafni, Inoue, Khairallah, and Habash}]{alhafni-etal-2023-advancements}
Bashar Alhafni, Go~Inoue, Christian Khairallah, and Nizar Habash. 2023.
\newblock \href {https://doi.org/10.18653/v1/2023.emnlp-main.396} {Advancements in {A}rabic grammatical error detection and correction: An empirical investigation}.
\newblock In \emph{Proceedings of the 2023 Conference on Empirical Methods in Natural Language Processing}, pages 6430--6448, Singapore. Association for Computational Linguistics.

\bibitem[{Antoun et~al.(2020)Antoun, Baly, and Hajj}]{antoun-etal-2020-arabert}
Wissam Antoun, Fady Baly, and Hazem Hajj. 2020.
\newblock \href {https://aclanthology.org/2020.osact-1.2/} {{A}ra{BERT}: Transformer-based model for {A}rabic language understanding}.
\newblock In \emph{Proceedings of the 4th Workshop on Open-Source Arabic Corpora and Processing Tools, with a Shared Task on Offensive Language Detection}, pages 9--15, Marseille, France. European Language Resource Association.

\bibitem[{Awasthi et~al.(2019)Awasthi, Sarawagi, Goyal, Ghosh, and Piratla}]{awasthi-etal-2019-parallel}
Abhijeet Awasthi, Sunita Sarawagi, Rasna Goyal, Sabyasachi Ghosh, and Vihari Piratla. 2019.
\newblock \href {https://doi.org/10.18653/v1/D19-1435} {Parallel iterative edit models for local sequence transduction}.
\newblock In \emph{Proceedings of the 2019 Conference on Empirical Methods in Natural Language Processing and the 9th International Joint Conference on Natural Language Processing (EMNLP-IJCNLP)}, pages 4260--4270, Hong Kong, China. Association for Computational Linguistics.

\bibitem[{Belkebir and Habash(2021)}]{belkebir-habash-2021-automatic}
Riadh Belkebir and Nizar Habash. 2021.
\newblock \href {https://doi.org/10.18653/v1/2021.conll-1.47} {Automatic error type annotation for {A}rabic}.
\newblock In \emph{Proceedings of the 25th Conference on Computational Natural Language Learning}, pages 596--606, Online. Association for Computational Linguistics.

\bibitem[{Bouamor et~al.(2018)Bouamor, Habash, Salameh, Zaghouani, Rambow, Abdulrahim, Obeid, Khalifa, Eryani, Erdmann, and Oflazer}]{bouamor-etal-2018-madar}
Houda Bouamor, Nizar Habash, Mohammad Salameh, Wajdi Zaghouani, Owen Rambow, Dana Abdulrahim, Ossama Obeid, Salam Khalifa, Fadhl Eryani, Alexander Erdmann, and Kemal Oflazer. 2018.
\newblock \href {https://aclanthology.org/L18-1535/} {The {MADAR} {A}rabic dialect corpus and lexicon}.
\newblock In \emph{Proceedings of the Eleventh International Conference on Language Resources and Evaluation ({LREC} 2018)}, Miyazaki, Japan. European Language Resources Association (ELRA).

\bibitem[{Bougares and Bouamor(2015)}]{bougares-bouamor-2015-ummu}
Fethi Bougares and Houda Bouamor. 2015.
\newblock \href {https://doi.org/10.18653/v1/W15-3221} {{UMMU}@{QALB}-2015 shared task: Character and word level {SMT} pipeline for automatic error correction of {A}rabic text}.
\newblock In \emph{Proceedings of the Second Workshop on {A}rabic Natural Language Processing}, pages 166--172, Beijing, China. Association for Computational Linguistics.

\bibitem[{Bryant et~al.(2019)Bryant, Felice, Andersen, and Briscoe}]{bryant-etal-2019-bea}
Christopher Bryant, Mariano Felice, {\O}istein~E. Andersen, and Ted Briscoe. 2019.
\newblock \href {https://doi.org/10.18653/v1/W19-4406} {The {BEA}-2019 shared task on grammatical error correction}.
\newblock In \emph{Proceedings of the Fourteenth Workshop on Innovative Use of NLP for Building Educational Applications}, pages 52--75, Florence, Italy. Association for Computational Linguistics.

\bibitem[{Bryant et~al.(2023)Bryant, Yuan, Qorib, Cao, Ng, and Briscoe}]{bryant-etal-2023-grammatical}
Christopher Bryant, Zheng Yuan, Muhammad~Reza Qorib, Hannan Cao, Hwee~Tou Ng, and Ted Briscoe. 2023.
\newblock \href {https://doi.org/10.1162/coli_a_00478} {Grammatical error correction: A survey of the state of the art}.
\newblock \emph{Computational Linguistics}, pages 643--701.

\bibitem[{Coyne et~al.(2023)Coyne, Sakaguchi, Galvan-Sosa, Zock, and Inui}]{coyne2023analyzingperformancegpt35gpt4}
Steven Coyne, Keisuke Sakaguchi, Diana Galvan-Sosa, Michael Zock, and Kentaro Inui. 2023.
\newblock \href {https://arxiv.org/abs/2303.14342} {Analyzing the performance of gpt-3.5 and gpt-4 in grammatical error correction}.
\newblock \emph{Preprint}, arXiv:2303.14342.

\bibitem[{Dahlmeier and Ng(2012)}]{dahlmeier-ng-2012-better}
Daniel Dahlmeier and Hwee~Tou Ng. 2012.
\newblock \href {https://aclanthology.org/N12-1067/} {Better evaluation for grammatical error correction}.
\newblock In \emph{Proceedings of the 2012 Conference of the North {A}merican Chapter of the Association for Computational Linguistics: Human Language Technologies}, pages 568--572, Montr{\'e}al, Canada. Association for Computational Linguistics.

\bibitem[{Davis et~al.(2024)Davis, Caines, Andersen, Taslimipoor, Yannakoudakis, Yuan, Bryant, Rei, and Buttery}]{davis-etal-2024-prompting}
Christopher Davis, Andrew Caines, {\O}istein~E. Andersen, Shiva Taslimipoor, Helen Yannakoudakis, Zheng Yuan, Christopher Bryant, Marek Rei, and Paula Buttery. 2024.
\newblock \href {https://doi.org/10.18653/v1/2024.findings-acl.711} {Prompting open-source and commercial language models for grammatical error correction of {E}nglish learner text}.
\newblock In \emph{Findings of the Association for Computational Linguistics: ACL 2024}, pages 11952--11967, Bangkok, Thailand. Association for Computational Linguistics.

\bibitem[{Devlin et~al.(2019)Devlin, Chang, Lee, and Toutanova}]{devlin-etal-2019-bert}
Jacob Devlin, Ming-Wei Chang, Kenton Lee, and Kristina Toutanova. 2019.
\newblock \href {https://doi.org/10.18653/v1/N19-1423} {{BERT}: Pre-training of deep bidirectional transformers for language understanding}.
\newblock In \emph{Proceedings of the 2019 Conference of the North {A}merican Chapter of the Association for Computational Linguistics: Human Language Technologies, Volume 1 (Long and Short Papers)}, pages 4171--4186, Minneapolis, Minnesota. Association for Computational Linguistics.

\bibitem[{Diab et~al.(2014)Diab, Al-Badrashiny, Aminian, Attia, Elfardy, Habash, Hawwari, Salloum, Dasigi, and Eskander}]{diab-etal-2014-tharwa}
Mona Diab, Mohamed Al-Badrashiny, Maryam Aminian, Mohammed Attia, Heba Elfardy, Nizar Habash, Abdelati Hawwari, Wael Salloum, Pradeep Dasigi, and Ramy Eskander. 2014.
\newblock \href {https://aclanthology.org/L14-1115/} {{T}harwa: A large scale dialectal {A}rabic - {S}tandard {A}rabic - {E}nglish lexicon}.
\newblock In \emph{Proceedings of the Ninth International Conference on Language Resources and Evaluation ({LREC}`14)}, pages 3782--3789, Reykjavik, Iceland. European Language Resources Association (ELRA).

\bibitem[{Eryani et~al.(2020)Eryani, Habash, Bouamor, and Khalifa}]{eryani-etal-2020-spelling}
Fadhl Eryani, Nizar Habash, Houda Bouamor, and Salam Khalifa. 2020.
\newblock \href {https://aclanthology.org/2020.lrec-1.508/} {A spelling correction corpus for multiple {A}rabic dialects}.
\newblock In \emph{Proceedings of the Twelfth Language Resources and Evaluation Conference}, pages 4130--4138, Marseille, France. European Language Resources Association.

\bibitem[{Eskander et~al.(2013)Eskander, Habash, Rambow, and Tomeh}]{eskander-etal-2013-processing}
Ramy Eskander, Nizar Habash, Owen Rambow, and Nadi Tomeh. 2013.
\newblock \href {https://aclanthology.org/N13-1066/} {Processing spontaneous orthography}.
\newblock In \emph{Proceedings of the 2013 Conference of the North {A}merican Chapter of the Association for Computational Linguistics: Human Language Technologies}, pages 585--595, Atlanta, Georgia. Association for Computational Linguistics.

\bibitem[{Fang et~al.(2023)Fang, Yang, Lan, Wong, Hu, Chao, and Zhang}]{fang2023chatgpt}
Tao Fang, Shu Yang, Kaixin Lan, Derek~F. Wong, Jinpeng Hu, Lidia~S. Chao, and Yue Zhang. 2023.
\newblock \href {https://arxiv.org/abs/2304.01746} {Is chatgpt a highly fluent grammatical error correction system? a comprehensive evaluation}.
\newblock \emph{Preprint}, arXiv:2304.01746.

\bibitem[{Farra et~al.(2014)Farra, Tomeh, Rozovskaya, and Habash}]{farra-etal-2014-generalized}
Noura Farra, Nadi Tomeh, Alla Rozovskaya, and Nizar Habash. 2014.
\newblock \href {https://doi.org/10.3115/v1/P14-2027} {Generalized character-level spelling error correction}.
\newblock In \emph{Proceedings of the 52nd Annual Meeting of the Association for Computational Linguistics (Volume 2: Short Papers)}, pages 161--167, Baltimore, Maryland. Association for Computational Linguistics.

\bibitem[{Ferguson(1959)}]{Ferguson:1959:diglossia}
Charles~F Ferguson. 1959.
\newblock {Diglossia}.
\newblock \emph{Word}, 15(2):325--340.

\bibitem[{Ghaddar et~al.(2022)Ghaddar, Wu, Bagga, Rashid, Bibi, Rezagholizadeh, Xing, Wang, Duan, Wang, Huai, Jiang, Liu, and Langlais}]{ghaddar-etal-2022-revisiting}
Abbas Ghaddar, Yimeng Wu, Sunyam Bagga, Ahmad Rashid, Khalil Bibi, Mehdi Rezagholizadeh, Chao Xing, Yasheng Wang, Xinyu Duan, Zhefeng Wang, Baoxing Huai, Xin Jiang, Qun Liu, and Phillippe Langlais. 2022.
\newblock \href {https://doi.org/10.18653/v1/2022.emnlp-main.205} {Revisiting pre-trained language models and their evaluation for {A}rabic natural language processing}.
\newblock In \emph{Proceedings of the 2022 Conference on Empirical Methods in Natural Language Processing}, pages 3135--3151, Abu Dhabi, United Arab Emirates. Association for Computational Linguistics.

\bibitem[{Grundkiewicz et~al.(2019)Grundkiewicz, Junczys-Dowmunt, and Heafield}]{grundkiewicz-etal-2019-neural}
Roman Grundkiewicz, Marcin Junczys-Dowmunt, and Kenneth Heafield. 2019.
\newblock \href {https://doi.org/10.18653/v1/W19-4427} {Neural grammatical error correction systems with unsupervised pre-training on synthetic data}.
\newblock In \emph{Proceedings of the Fourteenth Workshop on Innovative Use of NLP for Building Educational Applications}, pages 252--263, Florence, Italy. Association for Computational Linguistics.

\bibitem[{Habash et~al.(2012{\natexlab{a}})Habash, Diab, and Rambow}]{habash-etal-2012-conventional}
Nizar Habash, Mona Diab, and Owen Rambow. 2012{\natexlab{a}}.
\newblock \href {https://aclanthology.org/L12-1328/} {Conventional orthography for dialectal {A}rabic}.
\newblock In \emph{Proceedings of the Eighth International Conference on Language Resources and Evaluation ({LREC}`12)}, pages 711--718, Istanbul, Turkey. European Language Resources Association (ELRA).

\bibitem[{Habash et~al.(2018)Habash, Eryani, Khalifa, Rambow, Abdulrahim, Erdmann, Faraj, Zaghouani, Bouamor, Zalmout, Hassan, Al-Shargi, Alkhereyf, Abdulkareem, Eskander, Salameh, and Saddiki}]{habash-etal-2018-unified}
Nizar Habash, Fadhl Eryani, Salam Khalifa, Owen Rambow, Dana Abdulrahim, Alexander Erdmann, Reem Faraj, Wajdi Zaghouani, Houda Bouamor, Nasser Zalmout, Sara Hassan, Faisal Al-Shargi, Sakhar Alkhereyf, Basma Abdulkareem, Ramy Eskander, Mohammad Salameh, and Hind Saddiki. 2018.
\newblock \href {https://aclanthology.org/L18-1574/} {Unified guidelines and resources for {A}rabic dialect orthography}.
\newblock In \emph{Proceedings of the Eleventh International Conference on Language Resources and Evaluation ({LREC} 2018)}, Miyazaki, Japan. European Language Resources Association (ELRA).

\bibitem[{Habash et~al.(2012{\natexlab{b}})Habash, Eskander, and Hawwari}]{habash-etal-2012-morphological}
Nizar Habash, Ramy Eskander, and Abdelati Hawwari. 2012{\natexlab{b}}.
\newblock \href {https://aclanthology.org/W12-2301/} {A morphological analyzer for {E}gyptian {A}rabic}.
\newblock In \emph{Proceedings of the Twelfth Meeting of the Special Interest Group on Computational Morphology and Phonology}, pages 1--9, Montr{\'e}al, Canada. Association for Computational Linguistics.

\bibitem[{Habash and Palfreyman(2022)}]{habash-palfreyman-2022-zaebuc}
Nizar Habash and David Palfreyman. 2022.
\newblock \href {https://aclanthology.org/2022.lrec-1.9/} {{ZAEBUC}: An annotated {A}rabic-{E}nglish bilingual writer corpus}.
\newblock In \emph{Proceedings of the Thirteenth Language Resources and Evaluation Conference}, pages 79--88, Marseille, France. European Language Resources Association.

\bibitem[{Habash et~al.(2007)Habash, Soudi, and Buckwalter}]{Habash:2007:arabic-transliteration}
Nizar Habash, Abdelhadi Soudi, and Tim Buckwalter. 2007.
\newblock {On {{A}rabic} Transliteration}.
\newblock In A.~van~den Bosch and A.~Soudi, editors, \emph{{A}rabic Computational Morphology: Knowledge-based and Empirical Methods}, pages 15--22. Springer, Netherlands.

\bibitem[{Inoue et~al.(2021)Inoue, Alhafni, Baimukan, Bouamor, and Habash}]{inoue-etal-2021-interplay}
Go~Inoue, Bashar Alhafni, Nurpeiis Baimukan, Houda Bouamor, and Nizar Habash. 2021.
\newblock \href {https://aclanthology.org/2021.wanlp-1.10/} {The interplay of variant, size, and task type in {A}rabic pre-trained language models}.
\newblock In \emph{Proceedings of the Sixth Arabic Natural Language Processing Workshop}, pages 92--104, Kyiv, Ukraine (Virtual). Association for Computational Linguistics.

\bibitem[{Jarrar et~al.(2016)Jarrar, Habash, Alrimawi, Akra, and Zalmout}]{Jarrar:2016:curras}
Mustafa Jarrar, Nizar Habash, Faeq Alrimawi, Diyam Akra, and Nasser Zalmout. 2016.
\newblock {Curras: an annotated corpus for the Palestinian {A}rabic dialect}.
\newblock \emph{Language Resources and Evaluation}, pages 1--31.

\bibitem[{Junczys-Dowmunt et~al.(2018)Junczys-Dowmunt, Grundkiewicz, Guha, and Heafield}]{junczys-dowmunt-etal-2018-approaching}
Marcin Junczys-Dowmunt, Roman Grundkiewicz, Shubha Guha, and Kenneth Heafield. 2018.
\newblock \href {https://doi.org/10.18653/v1/N18-1055} {Approaching neural grammatical error correction as a low-resource machine translation task}.
\newblock In \emph{Proceedings of the 2018 Conference of the North {A}merican Chapter of the Association for Computational Linguistics: Human Language Technologies, Volume 1 (Long Papers)}, pages 595--606, New Orleans, Louisiana. Association for Computational Linguistics.

\bibitem[{Kaneko et~al.(2020)Kaneko, Mita, Kiyono, Suzuki, and Inui}]{kaneko-etal-2020-encoder}
Masahiro Kaneko, Masato Mita, Shun Kiyono, Jun Suzuki, and Kentaro Inui. 2020.
\newblock \href {https://doi.org/10.18653/v1/2020.acl-main.391} {Encoder-decoder models can benefit from pre-trained masked language models in grammatical error correction}.
\newblock In \emph{Proceedings of the 58th Annual Meeting of the Association for Computational Linguistics}, pages 4248--4254, Online. Association for Computational Linguistics.

\bibitem[{Kaneko and Okazaki(2023)}]{kaneko-okazaki-2023-reducing}
Masahiro Kaneko and Naoaki Okazaki. 2023.
\newblock \href {https://doi.org/10.18653/v1/2023.emnlp-main.619} {Reducing sequence length by predicting edit spans with large language models}.
\newblock In \emph{Proceedings of the 2023 Conference on Empirical Methods in Natural Language Processing}, pages 10017--10029, Singapore. Association for Computational Linguistics.

\bibitem[{Kaneko and Okazaki(2024)}]{kaneko-okazaki-2024-controlled}
Masahiro Kaneko and Naoaki Okazaki. 2024.
\newblock \href {https://aclanthology.org/2024.lrec-main.350/} {Controlled generation with prompt insertion for natural language explanations in grammatical error correction}.
\newblock In \emph{Proceedings of the 2024 Joint International Conference on Computational Linguistics, Language Resources and Evaluation (LREC-COLING 2024)}, pages 3955--3961, Torino, Italia. ELRA and ICCL.

\bibitem[{Katinskaia and Yangarber(2024)}]{katinskaia-yangarber-2024-gpt}
Anisia Katinskaia and Roman Yangarber. 2024.
\newblock \href {https://aclanthology.org/2024.lrec-main.692/} {{GPT}-3.5 for grammatical error correction}.
\newblock In \emph{Proceedings of the 2024 Joint International Conference on Computational Linguistics, Language Resources and Evaluation (LREC-COLING 2024)}, pages 7831--7843, Torino, Italia. ELRA and ICCL.

\bibitem[{Katsumata and Komachi(2020)}]{katsumata-komachi-2020-stronger}
Satoru Katsumata and Mamoru Komachi. 2020.
\newblock \href {https://doi.org/10.18653/v1/2020.aacl-main.83} {Stronger baselines for grammatical error correction using a pretrained encoder-decoder model}.
\newblock In \emph{Proceedings of the 1st Conference of the Asia-Pacific Chapter of the Association for Computational Linguistics and the 10th International Joint Conference on Natural Language Processing}, pages 827--832, Suzhou, China. Association for Computational Linguistics.

\bibitem[{Khalifa et~al.(2018)Khalifa, Habash, Eryani, Obeid, Abdulrahim, and Al~Kaabi}]{khalifa-etal-2018-morphologically}
Salam Khalifa, Nizar Habash, Fadhl Eryani, Ossama Obeid, Dana Abdulrahim, and Meera Al~Kaabi. 2018.
\newblock \href {https://aclanthology.org/L18-1607/} {A morphologically annotated corpus of emirati {A}rabic}.
\newblock In \emph{Proceedings of the Eleventh International Conference on Language Resources and Evaluation ({LREC} 2018)}, Miyazaki, Japan. European Language Resources Association (ELRA).

\bibitem[{Khalifa et~al.(2020)Khalifa, Zalmout, and Habash}]{khalifa-etal-2020-morphological}
Salam Khalifa, Nasser Zalmout, and Nizar Habash. 2020.
\newblock \href {https://aclanthology.org/2020.lrec-1.480/} {Morphological analysis and disambiguation for {G}ulf {A}rabic: The interplay between resources and methods}.
\newblock In \emph{Proceedings of the Twelfth Language Resources and Evaluation Conference}, pages 3895--3904, Marseille, France. European Language Resources Association.

\bibitem[{Kwon et~al.(2023)Kwon, Bhatia, Nagoudi, and Abdul-Mageed}]{kwon-etal-2023-beyond}
Sang Kwon, Gagan Bhatia, El~Moatez~Billah Nagoudi, and Muhammad Abdul-Mageed. 2023.
\newblock \href {https://doi.org/10.18653/v1/2023.arabicnlp-1.9} {Beyond {E}nglish: Evaluating {LLM}s for {A}rabic grammatical error correction}.
\newblock In \emph{Proceedings of ArabicNLP 2023}, pages 101--119, Singapore (Hybrid). Association for Computational Linguistics.

\bibitem[{{Levenshtein}(1966)}]{Levenshtein:1966:binary}
V.~I. {Levenshtein}. 1966.
\newblock {Binary Codes Capable of Correcting Deletions, Insertions and Reversals}.
\newblock \emph{Soviet Physics Doklady}, 10:707.

\bibitem[{Li et~al.(2022)Li, Parnow, and Zhao}]{Zuchao:2022}
Zuchao Li, Kevin Parnow, and Hai Zhao. 2022.
\newblock Incorporating rich syntax information in grammatical error correction.
\newblock \emph{Information Processing \& Management}, 59(3):102891.

\bibitem[{Loem et~al.(2023)Loem, Kaneko, Takase, and Okazaki}]{loem-etal-2023-exploring}
Mengsay Loem, Masahiro Kaneko, Sho Takase, and Naoaki Okazaki. 2023.
\newblock \href {https://doi.org/10.18653/v1/2023.bea-1.18} {Exploring effectiveness of {GPT}-3 in grammatical error correction: A study on performance and controllability in prompt-based methods}.
\newblock In \emph{Proceedings of the 18th Workshop on Innovative Use of NLP for Building Educational Applications (BEA 2023)}, pages 205--219, Toronto, Canada. Association for Computational Linguistics.

\bibitem[{Luhtaru et~al.(2024)Luhtaru, Korotkova, and Fishel}]{luhtaru-etal-2024-error}
Agnes Luhtaru, Elizaveta Korotkova, and Mark Fishel. 2024.
\newblock \href {https://aclanthology.org/2024.eacl-long.73/} {No error left behind: Multilingual grammatical error correction with pre-trained translation models}.
\newblock In \emph{Proceedings of the 18th Conference of the European Chapter of the Association for Computational Linguistics (Volume 1: Long Papers)}, pages 1209--1222, St. Julian{'}s, Malta. Association for Computational Linguistics.

\bibitem[{Maamouri et~al.(2014)Maamouri, Bies, Kulick, Ciul, Habash, and Eskander}]{ARZTB:2014}
Mohamed Maamouri, Ann Bies, Seth Kulick, Michael Ciul, Nizar Habash, and Ramy Eskander. 2014.
\newblock \href {http://www.lrec-conf.org/proceedings/lrec2014/pdf/1145_Paper.pdf} {Developing an egyptian arabic treebank: Impact of dialectal morphology on annotation and tool development}.
\newblock In \emph{Proceedings of the Ninth International Conference on Language Resources and Evaluation (LREC-2014)}. European Language Resources Association (ELRA).

\bibitem[{Magdy et~al.(2024)Magdy, Alwajih, Kwon, Abdel-Salam, and Abdul-Mageed}]{magdy-etal-2024-gazelle}
Samar~Mohamed Magdy, Fakhraddin Alwajih, Sang~Yun Kwon, Reem Abdel-Salam, and Muhammad Abdul-Mageed. 2024.
\newblock \href {https://doi.org/10.18653/v1/2024.findings-emnlp.941} {Gazelle: An instruction dataset for {A}rabic writing assistance}.
\newblock In \emph{Findings of the Association for Computational Linguistics: EMNLP 2024}, pages 16027--16054, Miami, Florida, USA. Association for Computational Linguistics.

\bibitem[{Mallinson et~al.(2022)Mallinson, Adamek, Malmi, and Severyn}]{mallinson-etal-2022-edit5}
Jonathan Mallinson, Jakub Adamek, Eric Malmi, and Aliaksei Severyn. 2022.
\newblock \href {https://doi.org/10.18653/v1/2022.findings-emnlp.156} {{E}di{T}5: Semi-autoregressive text editing with t5 warm-start}.
\newblock In \emph{Findings of the Association for Computational Linguistics: EMNLP 2022}, pages 2126--2138, Abu Dhabi, United Arab Emirates. Association for Computational Linguistics.

\bibitem[{Mallinson et~al.(2020)Mallinson, Severyn, Malmi, and Garrido}]{mallinson-etal-2020-felix}
Jonathan Mallinson, Aliaksei Severyn, Eric Malmi, and Guillermo Garrido. 2020.
\newblock \href {https://doi.org/10.18653/v1/2020.findings-emnlp.111} {{FELIX}: Flexible text editing through tagging and insertion}.
\newblock In \emph{Findings of the Association for Computational Linguistics: EMNLP 2020}, pages 1244--1255, Online. Association for Computational Linguistics.

\bibitem[{Malmi et~al.(2019)Malmi, Krause, Rothe, Mirylenka, and Severyn}]{malmi-etal-2019-encode}
Eric Malmi, Sebastian Krause, Sascha Rothe, Daniil Mirylenka, and Aliaksei Severyn. 2019.
\newblock \href {https://doi.org/10.18653/v1/D19-1510} {Encode, tag, realize: High-precision text editing}.
\newblock In \emph{Proceedings of the 2019 Conference on Empirical Methods in Natural Language Processing and the 9th International Joint Conference on Natural Language Processing (EMNLP-IJCNLP)}, pages 5054--5065, Hong Kong, China. Association for Computational Linguistics.

\bibitem[{Mesham et~al.(2023)Mesham, Bryant, Rei, and Yuan}]{mesham-etal-2023-extended}
Stuart Mesham, Christopher Bryant, Marek Rei, and Zheng Yuan. 2023.
\newblock \href {https://doi.org/10.18653/v1/2023.findings-eacl.119} {An extended sequence tagging vocabulary for grammatical error correction}.
\newblock In \emph{Findings of the Association for Computational Linguistics: EACL 2023}, pages 1608--1619, Dubrovnik, Croatia. Association for Computational Linguistics.

\bibitem[{Mita et~al.(2024)Mita, Sakaguchi, Hagiwara, Mizumoto, Suzuki, and Inui}]{mita-etal-2024-towards}
Masato Mita, Keisuke Sakaguchi, Masato Hagiwara, Tomoya Mizumoto, Jun Suzuki, and Kentaro Inui. 2024.
\newblock \href {https://aclanthology.org/2024.bea-1.21/} {Towards automated document revision: Grammatical error correction, fluency edits, and beyond}.
\newblock In \emph{Proceedings of the 19th Workshop on Innovative Use of NLP for Building Educational Applications (BEA 2024)}, pages 251--265, Mexico City, Mexico. Association for Computational Linguistics.

\bibitem[{Mohit et~al.(2014)Mohit, Rozovskaya, Habash, Zaghouani, and Obeid}]{mohit-etal-2014-first}
Behrang Mohit, Alla Rozovskaya, Nizar Habash, Wajdi Zaghouani, and Ossama Obeid. 2014.
\newblock \href {https://doi.org/10.3115/v1/W14-3605} {The first {QALB} shared task on automatic text correction for {A}rabic}.
\newblock In \emph{Proceedings of the {EMNLP} 2014 Workshop on {A}rabic Natural Language Processing ({ANLP})}, pages 39--47, Doha, Qatar. Association for Computational Linguistics.

\bibitem[{Nawar(2015)}]{nawar-2015-cufe}
Michael Nawar. 2015.
\newblock \href {https://doi.org/10.18653/v1/W15-3215} {{CUFE}@{QALB}-2015 shared task: {A}rabic error correction system}.
\newblock In \emph{Proceedings of the Second Workshop on {A}rabic Natural Language Processing}, pages 133--137, Beijing, China. Association for Computational Linguistics.

\bibitem[{Ng et~al.(2014)Ng, Wu, Briscoe, Hadiwinoto, Susanto, and Bryant}]{ng-etal-2014-conll}
Hwee~Tou Ng, Siew~Mei Wu, Ted Briscoe, Christian Hadiwinoto, Raymond~Hendy Susanto, and Christopher Bryant. 2014.
\newblock \href {https://doi.org/10.3115/v1/W14-1701} {The {C}o{NLL}-2014 shared task on grammatical error correction}.
\newblock In \emph{Proceedings of the Eighteenth Conference on Computational Natural Language Learning: Shared Task}, pages 1--14, Baltimore, Maryland. Association for Computational Linguistics.

\bibitem[{Ng et~al.(2013)Ng, Wu, Wu, Hadiwinoto, and Tetreault}]{ng-etal-2013-conll}
Hwee~Tou Ng, Siew~Mei Wu, Yuanbin Wu, Christian Hadiwinoto, and Joel Tetreault. 2013.
\newblock \href {https://aclanthology.org/W13-3601/} {The {C}o{NLL}-2013 shared task on grammatical error correction}.
\newblock In \emph{Proceedings of the Seventeenth Conference on Computational Natural Language Learning: Shared Task}, pages 1--12, Sofia, Bulgaria. Association for Computational Linguistics.

\bibitem[{Obeid et~al.(2022)Obeid, Inoue, and Habash}]{obeid-etal-2022-camelira}
Ossama Obeid, Go~Inoue, and Nizar Habash. 2022.
\newblock \href {https://doi.org/10.18653/v1/2022.emnlp-demos.32} {Camelira: An {A}rabic multi-dialect morphological disambiguator}.
\newblock In \emph{Proceedings of the 2022 Conference on Empirical Methods in Natural Language Processing: System Demonstrations}, pages 319--326, Abu Dhabi, UAE. Association for Computational Linguistics.

\bibitem[{Omelianchuk et~al.(2020)Omelianchuk, Atrasevych, Chernodub, and Skurzhanskyi}]{omelianchuk-etal-2020-gector}
Kostiantyn Omelianchuk, Vitaliy Atrasevych, Artem Chernodub, and Oleksandr Skurzhanskyi. 2020.
\newblock \href {https://doi.org/10.18653/v1/2020.bea-1.16} {{GECT}o{R} {--} grammatical error correction: Tag, not rewrite}.
\newblock In \emph{Proceedings of the Fifteenth Workshop on Innovative Use of NLP for Building Educational Applications}, pages 163--170, Seattle, WA, USA {\textrightarrow} Online. Association for Computational Linguistics.

\bibitem[{Omelianchuk et~al.(2024)Omelianchuk, Liubonko, Skurzhanskyi, Chernodub, Korniienko, and Samokhin}]{omelianchuk-etal-2024-pillars}
Kostiantyn Omelianchuk, Andrii Liubonko, Oleksandr Skurzhanskyi, Artem Chernodub, Oleksandr Korniienko, and Igor Samokhin. 2024.
\newblock \href {https://aclanthology.org/2024.bea-1.3/} {Pillars of grammatical error correction: Comprehensive inspection of contemporary approaches in the era of large language models}.
\newblock In \emph{Proceedings of the 19th Workshop on Innovative Use of NLP for Building Educational Applications (BEA 2024)}, pages 17--33, Mexico City, Mexico. Association for Computational Linguistics.

\bibitem[{OpenAI et~al.(2024)OpenAI, Achiam, Adler, Agarwal, Ahmad, Akkaya, Aleman, Almeida, Altenschmidt, Altman, Anadkat, Avila, Babuschkin, Balaji, Balcom, Baltescu, Bao, Bavarian, Belgum, Bello, Berdine, Bernadett-Shapiro, Berner, Bogdonoff, Boiko, Boyd, Brakman, Brockman, Brooks, Brundage, Button, Cai, Campbell, Cann, Carey, Carlson, Carmichael, Chan, Chang, Chantzis, Chen, Chen, Chen, Chen, Chen, Chess, Cho, Chu, Chung, Cummings, Currier, Dai, Decareaux, Degry, Deutsch, Deville, Dhar, Dohan, Dowling, Dunning, Ecoffet, Eleti, Eloundou, Farhi, Fedus, Felix, Fishman, Forte, Fulford, Gao, Georges, Gibson, Goel, Gogineni, Goh, Gontijo-Lopes, Gordon, Grafstein, Gray, Greene, Gross, Gu, Guo, Hallacy, Han, Harris, He, Heaton, Heidecke, Hesse, Hickey, Hickey, Hoeschele, Houghton, Hsu, Hu, Hu, Huizinga, Jain, Jain, Jang, Jiang, Jiang, Jin, Jin, Jomoto, Jonn, Jun, Kaftan, Łukasz Kaiser, Kamali, Kanitscheider, Keskar, Khan, Kilpatrick, Kim, Kim, Kim, Kirchner, Kiros, Knight, Kokotajlo, Łukasz Kondraciuk,
  Kondrich, Konstantinidis, Kosic, Krueger, Kuo, Lampe, Lan, Lee, Leike, Leung, Levy, Li, Lim, Lin, Lin, Litwin, Lopez, Lowe, Lue, Makanju, Malfacini, Manning, Markov, Markovski, Martin, Mayer, Mayne, McGrew, McKinney, McLeavey, McMillan, McNeil, Medina, Mehta, Menick, Metz, Mishchenko, Mishkin, Monaco, Morikawa, Mossing, Mu, Murati, Murk, Mély, Nair, Nakano, Nayak, Neelakantan, Ngo, Noh, Ouyang, O'Keefe, Pachocki, Paino, Palermo, Pantuliano, Parascandolo, Parish, Parparita, Passos, Pavlov, Peng, Perelman, de~Avila Belbute~Peres, Petrov, de~Oliveira~Pinto, Michael, Pokorny, Pokrass, Pong, Powell, Power, Power, Proehl, Puri, Radford, Rae, Ramesh, Raymond, Real, Rimbach, Ross, Rotsted, Roussez, Ryder, Saltarelli, Sanders, Santurkar, Sastry, Schmidt, Schnurr, Schulman, Selsam, Sheppard, Sherbakov, Shieh, Shoker, Shyam, Sidor, Sigler, Simens, Sitkin, Slama, Sohl, Sokolowsky, Song, Staudacher, Such, Summers, Sutskever, Tang, Tezak, Thompson, Tillet, Tootoonchian, Tseng, Tuggle, Turley, Tworek, Uribe, Vallone,
  Vijayvergiya, Voss, Wainwright, Wang, Wang, Wang, Ward, Wei, Weinmann, Welihinda, Welinder, Weng, Weng, Wiethoff, Willner, Winter, Wolrich, Wong, Workman, Wu, Wu, Wu, Xiao, Xu, Yoo, Yu, Yuan, Zaremba, Zellers, Zhang, Zhang, Zhao, Zheng, Zhuang, Zhuk, and Zoph}]{gpt4}
OpenAI, Josh Achiam, Steven Adler, Sandhini Agarwal, Lama Ahmad, Ilge Akkaya, Florencia~Leoni Aleman, Diogo Almeida, Janko Altenschmidt, Sam Altman, Shyamal Anadkat, Red Avila, Igor Babuschkin, Suchir Balaji, Valerie Balcom, Paul Baltescu, Haiming Bao, Mohammad Bavarian, Jeff Belgum, and 262 others. 2024.
\newblock \href {https://arxiv.org/abs/2303.08774} {Gpt-4 technical report}.
\newblock \emph{Preprint}, arXiv:2303.08774.

\bibitem[{Pasha et~al.(2014)Pasha, Al-Badrashiny, Diab, El~Kholy, Eskander, Habash, Pooleery, Rambow, and Roth}]{pasha-etal-2014-madamira}
Arfath Pasha, Mohamed Al-Badrashiny, Mona Diab, Ahmed El~Kholy, Ramy Eskander, Nizar Habash, Manoj Pooleery, Owen Rambow, and Ryan Roth. 2014.
\newblock \href {https://aclanthology.org/L14-1479/} {{MADAMIRA}: A fast, comprehensive tool for morphological analysis and disambiguation of {A}rabic}.
\newblock In \emph{Proceedings of the Ninth International Conference on Language Resources and Evaluation ({LREC}`14)}, pages 1094--1101, Reykjavik, Iceland. European Language Resources Association (ELRA).

\bibitem[{Qorib et~al.(2022)Qorib, Na, and Ng}]{qorib-etal-2022-frustratingly}
Muhammad~Reza Qorib, Seung-Hoon Na, and Hwee~Tou Ng. 2022.
\newblock \href {https://doi.org/10.18653/v1/2022.naacl-main.143} {Frustratingly easy system combination for grammatical error correction}.
\newblock In \emph{Proceedings of the 2022 Conference of the North American Chapter of the Association for Computational Linguistics: Human Language Technologies}, pages 1964--1974, Seattle, United States. Association for Computational Linguistics.

\bibitem[{Qorib and Ng(2023)}]{qorib-ng-2023-system}
Muhammad~Reza Qorib and Hwee~Tou Ng. 2023.
\newblock \href {https://doi.org/10.18653/v1/2023.emnlp-main.785} {System combination via quality estimation for grammatical error correction}.
\newblock In \emph{Proceedings of the 2023 Conference on Empirical Methods in Natural Language Processing}, pages 12746--12759, Singapore. Association for Computational Linguistics.

\bibitem[{Raheja et~al.(2024)Raheja, Alikaniotis, Kulkarni, Alhafni, and Kumar}]{raheja-etal-2024-medit}
Vipul Raheja, Dimitris Alikaniotis, Vivek Kulkarni, Bashar Alhafni, and Dhruv Kumar. 2024.
\newblock \href {https://doi.org/10.18653/v1/2024.naacl-long.56} {m{E}d{IT}: Multilingual text editing via instruction tuning}.
\newblock In \emph{Proceedings of the 2024 Conference of the North American Chapter of the Association for Computational Linguistics: Human Language Technologies (Volume 1: Long Papers)}, pages 979--1001, Mexico City, Mexico. Association for Computational Linguistics.

\bibitem[{Raheja et~al.(2023)Raheja, Kumar, Koo, and Kang}]{raheja-etal-2023-coedit}
Vipul Raheja, Dhruv Kumar, Ryan Koo, and Dongyeop Kang. 2023.
\newblock \href {https://doi.org/10.18653/v1/2023.findings-emnlp.350} {{C}o{E}d{IT}: Text editing by task-specific instruction tuning}.
\newblock In \emph{Findings of the Association for Computational Linguistics: EMNLP 2023}, pages 5274--5291, Singapore. Association for Computational Linguistics.

\bibitem[{Rothe et~al.(2021)Rothe, Mallinson, Malmi, Krause, and Severyn}]{rothe-etal-2021-simple}
Sascha Rothe, Jonathan Mallinson, Eric Malmi, Sebastian Krause, and Aliaksei Severyn. 2021.
\newblock \href {https://doi.org/10.18653/v1/2021.acl-short.89} {A simple recipe for multilingual grammatical error correction}.
\newblock In \emph{Proceedings of the 59th Annual Meeting of the Association for Computational Linguistics and the 11th International Joint Conference on Natural Language Processing (Volume 2: Short Papers)}, pages 702--707, Online. Association for Computational Linguistics.

\bibitem[{Rozovskaya et~al.(2015)Rozovskaya, Bouamor, Habash, Zaghouani, Obeid, and Mohit}]{rozovskaya-etal-2015-second}
Alla Rozovskaya, Houda Bouamor, Nizar Habash, Wajdi Zaghouani, Ossama Obeid, and Behrang Mohit. 2015.
\newblock \href {https://doi.org/10.18653/v1/W15-3204} {The second {QALB} shared task on automatic text correction for {A}rabic}.
\newblock In \emph{Proceedings of the Second Workshop on {A}rabic Natural Language Processing}, pages 26--35, Beijing, China. Association for Computational Linguistics.

\bibitem[{Rozovskaya et~al.(2014)Rozovskaya, Habash, Eskander, Farra, and Salloum}]{rozovskaya-etal-2014-columbia}
Alla Rozovskaya, Nizar Habash, Ramy Eskander, Noura Farra, and Wael Salloum. 2014.
\newblock \href {https://doi.org/10.3115/v1/W14-3622} {The {C}olumbia system in the {QALB}-2014 shared task on {A}rabic error correction}.
\newblock In \emph{Proceedings of the {EMNLP} 2014 Workshop on {A}rabic Natural Language Processing ({ANLP})}, pages 160--164, Doha, Qatar. Association for Computational Linguistics.

\bibitem[{Sengupta et~al.(2023)Sengupta, Sahu, Jia, Katipomu, Li, Koto, Marshall, Gosal, Liu, Chen, Afzal, Kamboj, Pandit, Pal, Pradhan, Mujahid, Baali, Han, Bsharat, Aji, Shen, Liu, Vassilieva, Hestness, Hock, Feldman, Lee, Jackson, Ren, Nakov, Baldwin, and Xing}]{sengupta2023jais}
Neha Sengupta, Sunil~Kumar Sahu, Bokang Jia, Satheesh Katipomu, Haonan Li, Fajri Koto, William Marshall, Gurpreet Gosal, Cynthia Liu, Zhiming Chen, Osama~Mohammed Afzal, Samta Kamboj, Onkar Pandit, Rahul Pal, Lalit Pradhan, Zain~Muhammad Mujahid, Massa Baali, Xudong Han, Sondos~Mahmoud Bsharat, and 13 others. 2023.
\newblock \href {https://arxiv.org/abs/2308.16149} {Jais and jais-chat: Arabic-centric foundation and instruction-tuned open generative large language models}.
\newblock \emph{Preprint}, arXiv:2308.16149.

\bibitem[{Solyman et~al.(2022)Solyman, Wang, Tao, Elhag, Zhang, and Mahmoud}]{solyman-etal-2022-automatic}
Aiman Solyman, Zhenyu Wang, Qian Tao, Arafat Abdulgader~Mohammed Elhag, Rui Zhang, and Zeinab Mahmoud. 2022.
\newblock \href {https://doi.org/10.1016/j.knosys.2022.108180} {Automatic {Arabic} grammatical error correction based on expectation-maximization routing and target-bidirectional agreement}.
\newblock \emph{Knowledge-Based Systems}, 241:108180.

\bibitem[{Solyman et~al.(2023)Solyman, Zappatore, Zhenyu, Mahmoud, Alfatemi, Ibrahim, and Gabralla}]{solyman-etal-2023-optimizing}
Aiman Solyman, Marco Zappatore, Wang Zhenyu, Zeinab Mahmoud, Ali Alfatemi, Ashraf~Osman Ibrahim, and Lubna~Abdelkareim Gabralla. 2023.
\newblock \href {https://doi.org/10.1016/j.jksuci.2023.101572} {Optimizing the impact of data augmentation for low-resource grammatical error correction}.
\newblock \emph{Journal of King Saud University - Computer and Information Sciences}, 35(6):101572.

\bibitem[{Solyman et~al.(2021)Solyman, Zhenyu, Qian, Elhag, Toseef, and Aleibeid}]{solyman-etal-2021-synthetic}
Aiman Solyman, Wang Zhenyu, Tao Qian, Arafat Abdulgader~Mohammed Elhag, Muhammad Toseef, and Zeinab Aleibeid. 2021.
\newblock \href {https://doi.org/10.1016/j.eij.2020.12.001} {Synthetic data with neural machine translation for automatic correction in {Arabic} grammar}.
\newblock \emph{Egyptian Informatics Journal}, 22(3):303--315.

\bibitem[{Stahlberg and Kumar(2020)}]{stahlberg-kumar-2020-seq2edits}
Felix Stahlberg and Shankar Kumar. 2020.
\newblock \href {https://doi.org/10.18653/v1/2020.emnlp-main.418} {{S}eq2{E}dits: Sequence transduction using span-level edit operations}.
\newblock In \emph{Proceedings of the 2020 Conference on Empirical Methods in Natural Language Processing (EMNLP)}, pages 5147--5159, Online. Association for Computational Linguistics.

\bibitem[{Stahlberg and Kumar(2021)}]{stahlberg-kumar-2021-synthetic}
Felix Stahlberg and Shankar Kumar. 2021.
\newblock \href {https://aclanthology.org/2021.bea-1.4/} {Synthetic data generation for grammatical error correction with tagged corruption models}.
\newblock In \emph{Proceedings of the 16th Workshop on Innovative Use of NLP for Building Educational Applications}, pages 37--47, Online. Association for Computational Linguistics.

\bibitem[{Stahlberg and Kumar(2024)}]{stahlberg-kumar-2024-synthetic}
Felix Stahlberg and Shankar Kumar. 2024.
\newblock \href {https://aclanthology.org/2024.bea-1.2/} {Synthetic data generation for low-resource grammatical error correction with tagged corruption models}.
\newblock In \emph{Proceedings of the 19th Workshop on Innovative Use of NLP for Building Educational Applications (BEA 2024)}, pages 11--16, Mexico City, Mexico. Association for Computational Linguistics.

\bibitem[{Straka et~al.(2021)Straka, N{\'a}plava, and Strakov{\'a}}]{straka-etal-2021-character}
Milan Straka, Jakub N{\'a}plava, and Jana Strakov{\'a}. 2021.
\newblock \href {https://doi.org/10.18653/v1/2021.wnut-1.46} {Character transformations for non-autoregressive {GEC} tagging}.
\newblock In \emph{Proceedings of the Seventh Workshop on Noisy User-generated Text (W-NUT 2021)}, pages 417--422, Online. Association for Computational Linguistics.

\bibitem[{Tarnavskyi et~al.(2022)Tarnavskyi, Chernodub, and Omelianchuk}]{tarnavskyi-etal-2022-ensembling}
Maksym Tarnavskyi, Artem Chernodub, and Kostiantyn Omelianchuk. 2022.
\newblock \href {https://doi.org/10.18653/v1/2022.acl-long.266} {Ensembling and knowledge distilling of large sequence taggers for grammatical error correction}.
\newblock In \emph{Proceedings of the 60th Annual Meeting of the Association for Computational Linguistics (Volume 1: Long Papers)}, pages 3842--3852, Dublin, Ireland. Association for Computational Linguistics.

\bibitem[{Team et~al.(2025)Team, Abbas, Ahmad, Alam, Altinisik, Asgari, Boshmaf, Boughorbel, Chawla, Chowdhury, Dalvi, Darwish, Durrani, Elfeky, Elmagarmid, Eltabakh, Fatehkia, Fragkopoulos, Hasanain, Hawasly, Husaini, Jung, Lucas, Magdy, Messaoud, Mohamed, Mohiuddin, Mousi, Mubarak, Musleh, Naeem, Ouzzani, Popovic, Sadeghi, Sencar, Shinoy, Sinan, Zhang, Ali, Kheir, Ma, and Ruan}]{fanar2025}
Fanar Team, Ummar Abbas, Mohammad~Shahmeer Ahmad, Firoj Alam, Enes Altinisik, Ehsannedin Asgari, Yazan Boshmaf, Sabri Boughorbel, Sanjay Chawla, Shammur Chowdhury, Fahim Dalvi, Kareem Darwish, Nadir Durrani, Mohamed Elfeky, Ahmed Elmagarmid, Mohamed Eltabakh, Masoomali Fatehkia, Anastasios Fragkopoulos, Maram Hasanain, and 23 others. 2025.
\newblock \href {https://arxiv.org/abs/2501.13944} {Fanar: An arabic-centric multimodal generative ai platform}.
\newblock \emph{Preprint}, arXiv:2501.13944.

\bibitem[{Wan et~al.(2020)Wan, Wan, and Wang}]{wan-etal-2020-improving}
Zhaohong Wan, Xiaojun Wan, and Wenguang Wang. 2020.
\newblock \href {https://doi.org/10.18653/v1/2020.coling-main.200} {Improving grammatical error correction with data augmentation by editing latent representation}.
\newblock In \emph{Proceedings of the 28th International Conference on Computational Linguistics}, pages 2202--2212, Barcelona, Spain (Online). International Committee on Computational Linguistics.

\bibitem[{Watson et~al.(2018)Watson, Zalmout, and Habash}]{watson-etal-2018-utilizing}
Daniel Watson, Nasser Zalmout, and Nizar Habash. 2018.
\newblock \href {https://doi.org/10.18653/v1/D18-1097} {Utilizing character and word embeddings for text normalization with sequence-to-sequence models}.
\newblock In \emph{Proceedings of the 2018 Conference on Empirical Methods in Natural Language Processing}, pages 837--843, Brussels, Belgium. Association for Computational Linguistics.

\bibitem[{Wolf et~al.(2020)Wolf, Debut, Sanh, Chaumond, Delangue, Moi, Cistac, Rault, Louf, Funtowicz, Davison, Shleifer, von Platen, Ma, Jernite, Plu, Xu, Le~Scao, Gugger, Drame, Lhoest, and Rush}]{wolf-etal-2020-transformers}
Thomas Wolf, Lysandre Debut, Victor Sanh, Julien Chaumond, Clement Delangue, Anthony Moi, Pierric Cistac, Tim Rault, Remi Louf, Morgan Funtowicz, Joe Davison, Sam Shleifer, Patrick von Platen, Clara Ma, Yacine Jernite, Julien Plu, Canwen Xu, Teven Le~Scao, Sylvain Gugger, and 3 others. 2020.
\newblock \href {https://doi.org/10.18653/v1/2020.emnlp-demos.6} {Transformers: State-of-the-art natural language processing}.
\newblock In \emph{Proceedings of the 2020 Conference on Empirical Methods in Natural Language Processing: System Demonstrations}, pages 38--45, Online. Association for Computational Linguistics.

\bibitem[{Wu et~al.(2023)Wu, Wang, Wan, Jiao, and Lyu}]{wu2023chatgptgrammarlyevaluatingchatgpt}
Haoran Wu, Wenxuan Wang, Yuxuan Wan, Wenxiang Jiao, and Michael Lyu. 2023.
\newblock \href {https://arxiv.org/abs/2303.13648} {Chatgpt or grammarly? evaluating chatgpt on grammatical error correction benchmark}.
\newblock \emph{Preprint}, arXiv:2303.13648.

\bibitem[{Yuan and Bryant(2021)}]{yuan-bryant-2021-document}
Zheng Yuan and Christopher Bryant. 2021.
\newblock \href {https://aclanthology.org/2021.bea-1.8/} {Document-level grammatical error correction}.
\newblock In \emph{Proceedings of the 16th Workshop on Innovative Use of NLP for Building Educational Applications}, pages 75--84, Online. Association for Computational Linguistics.

\bibitem[{Yuan et~al.(2019)Yuan, Stahlberg, Rei, Byrne, and Yannakoudakis}]{yuan-etal-2019-neural}
Zheng Yuan, Felix Stahlberg, Marek Rei, Bill Byrne, and Helen Yannakoudakis. 2019.
\newblock \href {https://doi.org/10.18653/v1/W19-4424} {Neural and {FST}-based approaches to grammatical error correction}.
\newblock In \emph{Proceedings of the Fourteenth Workshop on Innovative Use of NLP for Building Educational Applications}, pages 228--239, Florence, Italy. Association for Computational Linguistics.

\bibitem[{Yuan et~al.(2021)Yuan, Taslimipoor, Davis, and Bryant}]{yuan-etal-2021-multi}
Zheng Yuan, Shiva Taslimipoor, Christopher Davis, and Christopher Bryant. 2021.
\newblock \href {https://doi.org/10.18653/v1/2021.emnlp-main.687} {{M}ulti-class grammatical error detection for correction: {A} tale of two systems}.
\newblock In \emph{Proceedings of the 2021 Conference on Empirical Methods in Natural Language Processing}, pages 8722--8736, Online and Punta Cana, Dominican Republic. Association for Computational Linguistics.

\bibitem[{Zaghouani et~al.(2015)Zaghouani, Habash, Bouamor, Rozovskaya, Mohit, Heider, and Oflazer}]{zaghouani-etal-2015-correction}
Wajdi Zaghouani, Nizar Habash, Houda Bouamor, Alla Rozovskaya, Behrang Mohit, Abeer Heider, and Kemal Oflazer. 2015.
\newblock \href {https://doi.org/10.3115/v1/W15-1614} {Correction annotation for non-native {A}rabic texts: Guidelines and corpus}.
\newblock In \emph{Proceedings of the 9th Linguistic Annotation Workshop}, pages 129--139, Denver, Colorado, USA. Association for Computational Linguistics.

\bibitem[{Zaghouani et~al.(2014)Zaghouani, Mohit, Habash, Obeid, Tomeh, Rozovskaya, Farra, Alkuhlani, and Oflazer}]{zaghouani-etal-2014-large}
Wajdi Zaghouani, Behrang Mohit, Nizar Habash, Ossama Obeid, Nadi Tomeh, Alla Rozovskaya, Noura Farra, Sarah Alkuhlani, and Kemal Oflazer. 2014.
\newblock \href {https://aclanthology.org/L14-1721/} {Large scale {A}rabic error annotation: Guidelines and framework}.
\newblock In \emph{Proceedings of the Ninth International Conference on Language Resources and Evaluation ({LREC}`14)}, Reykjavik, Iceland. European Language Resources Association (ELRA).

\bibitem[{Zalmout et~al.(2018)Zalmout, Erdmann, and Habash}]{zalmout-etal-2018-noise}
Nasser Zalmout, Alexander Erdmann, and Nizar Habash. 2018.
\newblock \href {https://doi.org/10.18653/v1/N18-1087} {Noise-robust morphological disambiguation for dialectal {A}rabic}.
\newblock In \emph{Proceedings of the 2018 Conference of the North {A}merican Chapter of the Association for Computational Linguistics: Human Language Technologies, Volume 1 (Long Papers)}, pages 953--964, New Orleans, Louisiana. Association for Computational Linguistics.

\bibitem[{Zalmout and Habash(2020)}]{zalmout-habash-2020-joint}
Nasser Zalmout and Nizar Habash. 2020.
\newblock \href {https://doi.org/10.18653/v1/2020.acl-main.736} {Joint diacritization, lemmatization, normalization, and fine-grained morphological tagging}.
\newblock In \emph{Proceedings of the 58th Annual Meeting of the Association for Computational Linguistics}, pages 8297--8307, Online. Association for Computational Linguistics.

\bibitem[{Zhang et~al.(2023)Zhang, Zhang, Cui, and Fu}]{zhang-etal-2023-non}
Yu~Zhang, Yue Zhang, Leyang Cui, and Guohong Fu. 2023.
\newblock \href {https://doi.org/10.18653/v1/2023.emnlp-main.437} {Non-autoregressive text editing with copy-aware latent alignments}.
\newblock In \emph{Proceedings of the 2023 Conference on Empirical Methods in Natural Language Processing}, pages 7075--7085, Singapore. Association for Computational Linguistics.

\bibitem[{Zhang et~al.(2022)Zhang, Zhang, Li, Bao, Li, and Zhang}]{zhang-etal-2022-syngec}
Yue Zhang, Bo~Zhang, Zhenghua Li, Zuyi Bao, Chen Li, and Min Zhang. 2022.
\newblock \href {https://doi.org/10.18653/v1/2022.emnlp-main.162} {{S}yn{GEC}: Syntax-enhanced grammatical error correction with a tailored {GEC}-oriented parser}.
\newblock In \emph{Proceedings of the 2022 Conference on Empirical Methods in Natural Language Processing}, pages 2518--2531, Abu Dhabi, United Arab Emirates. Association for Computational Linguistics.

\bibitem[{Zhao et~al.(2019)Zhao, Wang, Shen, Jia, and Liu}]{zhao-etal-2019-improving}
Wei Zhao, Liang Wang, Kewei Shen, Ruoyu Jia, and Jingming Liu. 2019.
\newblock \href {https://doi.org/10.18653/v1/N19-1014} {Improving grammatical error correction via pre-training a copy-augmented architecture with unlabeled data}.
\newblock In \emph{Proceedings of the 2019 Conference of the North {A}merican Chapter of the Association for Computational Linguistics: Human Language Technologies, Volume 1 (Long and Short Papers)}, pages 156--165, Minneapolis, Minnesota. Association for Computational Linguistics.

\bibitem[{Zhou et~al.(2023)Zhou, Liu, Li, Zhang, Zhang, Li, Zhang, and Huang}]{zhou-etal-2023-improving-seq2seq}
Houquan Zhou, Yumeng Liu, Zhenghua Li, Min Zhang, Bo~Zhang, Chen Li, Ji~Zhang, and Fei Huang. 2023.
\newblock \href {https://doi.org/10.18653/v1/2023.findings-emnlp.495} {Improving {S}eq2{S}eq grammatical error correction via decoding interventions}.
\newblock In \emph{Findings of the Association for Computational Linguistics: EMNLP 2023}, pages 7393--7405, Singapore. Association for Computational Linguistics.

\end{thebibliography}

\appendix
\newpage 
\section{Hyperparameters}
\label{sec:hyperparams}
We use Hugging Face's Transformers to build our edit taggers. Models trained on QALB-2014 or MADAR CODA are fine-tuned for 10 epochs using a learning rate of 5e-5, a batch size of 32, a maximum sequence length of 512, and a seed of 42 on a single A100 GPU. For models trained on QALB-2014 with the tenfold upsampled ZAEBUC, we use the same hyperparameters but run training for 15 epochs. At the end of fine-tuning, we pick the best checkpoint based on the performance on the Dev sets by using the M\textsuperscript{2} scorer.

% \begin{itemize}
%     \item Prompts
%     \item Hyperparameters
%     \item Full results using different BERT Models
%     \item Full results using LLMs with different settings
% \end{itemize}

\newpage
\onecolumn
\section{Edit Coverage}
\label{sec:appendix-edits-coverage}

% https://docs.google.com/spreadsheets/d/1LPT-S8FZBSbl85r02KGMECYCsPt9POgQSjuTLpdsdWc/edit?gid=548125526#gid=548125526

\begin{table*}[ht]
\setlength{\tabcolsep}{3pt}
\small
\centering
\begin{tabular}{lccc|ccc|ccc|ccc}
\toprule

 \multicolumn{4}{c}{} & \multicolumn{3}{c}{\textbf{QALB-2014}} &  \multicolumn{3}{c}{\textbf{ZAEBUC}} & \multicolumn{3}{c}{\textbf{MADAR CODA}}  \\
 
\bf Input & \bf Comp.  & \bf Subset  & \bf Prune  & \bf Edits & \bf OOV\%  & \bf F\textsubscript{0.5}  & \bf Edits & \bf OOV\%  & \bf F\textsubscript{0.5}  & \bf Edits & \bf OOV\%  & \bf F\textsubscript{0.5} \\\hline

Word & \XSolidBrush & All & - & 16,221  & 1.00\% & 98.4 & 1,097 & 2.94\% & 96.2 & 1,228 & 1.52\% & 98.0 \\
Subword & \XSolidBrush & All & - & 9,060  &  0.36\% & 98.7 & 905 & 1.85\% & 96.5 & 677 & 0.55\% & 98.1 \\\hline

Word & \Checkmark & All & - & 10,410 &  1.00\% & 98.4 & 687 & 2.94\% & 96.2 & 741 & 1.52\% & 98.0 \\
Subword & \Checkmark & All & - &6,170  &  0.36\% & 98.7 & 563 & 1.85\% & 96.5 & 454 & 0.55\% & 98.1 \\\hline

Subword & \Checkmark & NoPnx & - & 4,799  &  0.27\% & 98.8& 498 & 1.74\% & 96.2 & - & - & -  \\
Subword & \Checkmark & Pnx & - & 160 & 0.01\% & 99.4 & 23  & 0.06\% & 99.9 &  - & - & -  \\\hline

Subword & \Checkmark & All & 10 & 683 & 0.75\% & 98.1 & 58  & 3.71\% & 93.9 & 84 & 1.33\% & 96.2 \\
Subword & \Checkmark & All & 20 & 442 & 1.02\% & 97.7& 35  & 4.67\% & 92.6 & 52 & 2.02\% & 94.1 \\
Subword & \Checkmark & All & 30 & 329 & 1.24\% & 97.4 & 27  & 5.26\% & 91.8 & 45 & 2.28\% & 93.4 \\\hline

Subword & \Checkmark & NoPnx & 10 & 520 & 0.56\% & 98.2 & 52  & 3.39\% & 93.7 &  - & - & - \\
Subword & \Checkmark & NoPnx & 20 & 335 & 0.75\% & 97.8 & 30  & 4.31\% & 92.3 &  - & - & - \\
Subword & \Checkmark & NoPnx & 30 & 250 & 0.92\% & 97.5 & 22 & 4.90\% & 91.4 &  - & - & -  \\\hline

Subword & \Checkmark & Pnx & 10 & 48  & 0.02\% &  99.4 & 6 & 0.11\% & 99.9 & - & - & -  \\
Subword & \Checkmark & Pnx & 20 & 35 & 0.05\% & 99.4 & 6 & 0.11\% & 99.9 &  - & - & -  \\
Subword & \Checkmark & Pnx & 30 & 29 &  0.05\% & 99.3 & 6 & 0.11\% & 99.9 &  - & - & -  \\
\bottomrule

\end{tabular}
\caption{Edit statistics on QALB-2014, ZAEBUC and MADAR CODA. \textbf{Input} is the input unit (word or subword). \textbf{Comp.} indicates whether the edit is compressed. \textbf{Subset} specifies whether the edits capture all errors, punctuation-only errors (Pnx), or non-punctuation errors (NoPnx). \textbf{Edits} represents the total number of unique edits in the training set of each dataset. \textbf{OOV\%} is the percentage of out-of-vocabulary edits (non-unique) in the Dev set of each dataset.}

\label{tab:edit-coverage-all}
\end{table*}

\newpage
\onecolumn
\section{Edit Tagging Results}
\label{sec:appendix-bert-res}

\begin{table*}[h]
\setlength{\tabcolsep}{3pt}
\small
\centering
\begin{tabular}{llccc|cccc|cccc|cccc}
\toprule

\multicolumn{5}{c}{} & \multicolumn{4}{c}{\bf QALB-2014} & \multicolumn{4}{c}{\bf ZAEBUC} & \multicolumn{4}{c}{\bf MADAR CODA }\\

\textbf{Model} &\textbf{Input} & \textbf{Comp.}  & \textbf{Subset} & \textbf{Prune} & {\bf P} & {\bf R} & {\bf F\textsubscript{1}} & {\bf F\textsubscript{0.5}} & {\bf P} & {\bf R} & {\bf F\textsubscript{1}} & {\bf F\textsubscript{0.5}} & {\bf P} & {\bf R} & {\bf F\textsubscript{1}} & {\bf F\textsubscript{0.5}} \\
\hline

AraBERTv02 & Word & {\XSolidBrush} & All & - & 
81.0 &  64.3 &  71.7 &  77.0 &  84.8 &  69.5 &  76.4 &  81.2 &  87.9 &  66.5 &  75.7 &  82.6 \\
AraBERTv02 & Subword & {\XSolidBrush} & All & - & 81.0 & 67.8 &  73.8 &  77.9 &  84.4 &  71.3 &  77.3 &  81.4 &  87.6 &  76.8 &  81.9 &  85.2 \\
AraBERTv02 & Word & {\CheckmarkBold}  & All & - & 80.8 &   66.6 &  73.0 &  77.5 &  83.8 &  71.4 &  77.1 &  81.0 &  85.6 &  76.9 &  81.0 &  83.7 \\
AraBERTv02 & Subword & {\CheckmarkBold}  & All & - & 81.1 &   \textbf{69.1} &  74.6 &  78.4 &  84.3 &  \textbf{72.9} &  78.2 &  81.7 &  86.9 &  \textbf{79.2} &  \textbf{82.9 }&  85.2 \\\hline

ARBERTv2 & Word & {\XSolidBrush} & All & - & 78.9 &  57.7 &  66.7 &  73.5 &  82.2 &  54.8 &  65.8 &  74.8 &  86.4 &  61.0 &  71.5 &  79.8 \\
{ARBERTv2} & Subword & {\XSolidBrush} & All & - & 78.7 &    60.8 &  68.6 &  74.3 &  79.7 &  58.1 &  67.2 &  74.2 &  84.5 &  69.0 &  76.0 &  80.8 \\
{ARBERTv2} & Word & {\CheckmarkBold} & All & - & 77.8 &   61.4 &  68.6 &  73.8 &  80.7 &  62.8 &  70.6 &  76.3 &  81.8 &  68.4 &  74.5 &  78.7 \\
{ARBERTv2} & Subword & {\CheckmarkBold} & All & - & 78.6 &   60.0 &  68.0 &  74.0 &  82.7 &  62.1 &  70.9 &  77.5 &  84.2 &  70.8 &  77.0 &  81.2 \\\hline

CAMeLBERT & Word & {\XSolidBrush} & All & - & 81.2 &   61.5 &  70.0 &  76.3 &  84.6 &  66.4 &  74.4 &  80.2 &  88.3 &  66.4 &  75.8 &  82.8 \\
CAMeLBERT & Subword & {\XSolidBrush} & All & - & 80.4 &   65.2 &  72.0 &  76.9 &  83.5 &  69.3 &  75.8 &  80.2 &  87.1 &  76.8 &  81.6 &  84.8 \\
CAMeLBERT & Word & {\CheckmarkBold}  & All & - & 79.9 & 65.4 &  71.9 &  76.5 &  84.2 &  69.3 &  76.0 &  80.7 &  85.6 &  76.0 &  80.6 &  83.5 \\
CAMeLBERT & Subword & {\CheckmarkBold}  & All & - & 80.6 & 67.4 &  73.4 &  77.6 &  84.6 &  70.8 &  77.1 &  81.4 &  87.0 &  78.8 &  82.7 &  85.2 \\\hline

AraBERTv02 & Subword & {\CheckmarkBold} & All & 10 & \textbf{81.8} & 68.8 & \textbf{74.7}&  \textbf{78.8} &  84.5 &  71.9 &  77.7 &  81.6 &  \textbf{89.1} &  75.5 &  81.7 &  \textbf{86.0} \\
AraBERTv02 & Subword & {\CheckmarkBold} &  All & 20 &  81.4 & 68.6 &  74.4 &  78.5 &  85.3 &  72.0 &  78.1 &  82.2 &  87.7 &  73.1 &  79.8 &  84.4 \\
AraBERTv02 & Subword & {\CheckmarkBold} & All & 30 & 81.6 & 68.1 &  74.3 &  78.5 &  \textbf{85.8} &  72.3 &  \textbf{78.4} &  \textbf{82.7} &  88.3 &  72.1 &  79.4 &  84.5 \\\hline

{CAMeLBERT} & Subword & {\CheckmarkBold}  & All & 10 & 81.2 &   67.4 &  73.7 &  78.0 &  85.1 &  71.0 &  77.4 &  81.8 &  88.4 &  76.3 &  81.9 &  85.7 \\
{CAMeLBERT} & Subword & {\CheckmarkBold}  & All & 20 & 81.3 &   66.7 &  73.3 &  77.9 &  84.4 &  70.1 &  76.6 &  81.1 &  88.2 &  72.6 &  79.6 &  84.6 \\
{CAMeLBERT} & Subword & {\CheckmarkBold}  & All & 30 & 81.1 &   67.5 &  73.7 &  77.9 &  84.7 &  70.0 &  76.6 &  81.3 &  88.7 &  71.3 &  79.1 &  84.6 \\\hline\hline

AraBERTv02 & Subword & {\CheckmarkBold} & NoPnx & - & 88.3 & 77.7 & 82.6 & 85.9 & 87.2 & \underline{77.0} & 81.8 & 85.0 &  - & - &  - &  - \\
AraBERTv02 & Subword & {\CheckmarkBold} & NoPnx & 10 & 88.8 & \underline{78.1} & \underline{83.1} & 86.4 & 87.6 & 76.1 & 81.4 & 85.0 &  - & - &  - &  - \\
AraBERTv02 & Subword & {\CheckmarkBold} &  NoPnx & 20 &  89.0 & 77.8 & 83.0 & 86.5 & 87.9 & 75.8 & 81.4 & 85.1 &  - & - &  - &  - \\
AraBERTv02 & Subword & {\CheckmarkBold} & NoPnx & 30 &  \underline{89.4} & 77.5 & 83.0 & \underline{86.7} & \underline{88.1} & 76.8 & \underline{82.1} & \underline{85.6} &  - & - &  - &  - \\\hline

AraBERTv02 & Subword & {\CheckmarkBold} & Pnx & - & 90.6 & 83.0 & 86.6 & \underline{89.0} & 96.8 & 94.0 & \underline{95.4} & 96.2 &  - & - &  - &  - \\
AraBERTv02 & Subword & {\CheckmarkBold} & Pnx & 10 & 89.5 & \underline{83.6} & 86.5 & 88.3 & \underline{96.9} & 93.8 & 95.3 & \underline{96.3} &  - & - &  - &  - \\
AraBERTv02 & Subword & {\CheckmarkBold} &  Pnx & 20 &  \underline{90.7} & 82.8 & 86.5 & \underline{89.0} & 96.7 & 93.6 & 95.1 & 96.1 &  - & - &  - &  - \\
AraBERTv02 & Subword & {\CheckmarkBold} & Pnx & 30 & 90.1 & \underline{83.6} & \underline{86.7} & 88.7 & 96.5 & \underline{94.0} & 95.2 & 96.0 &  - & - &  - &  - \\

\bottomrule

\end{tabular}
\caption{MSA and DA GEC results on the Dev sets of QALB-2014, ZAEBUC, and MADAR CODA. \textbf{Input} is the input unit (word or subword). \textbf{Comp.} indicates whether the edit is compressed. \textbf{Subset} specifies whether the edits capture all errors, punctuation-only errors (Pnx), or non-punctuation errors (NoPnx). NoPnx models are evaluated after removing punctuation, while Pnx models are evaluated on a version of the Dev set where all non-punctuation errors are corrected.  Pruning experiments were conducted using the top two models (AraBERTv02 and CAMeLBERT), while punctuation segregation experiments used the best model (AraBERTv02). Best \textbf{All} results are in bold; best \textbf{NoPnx} and \textbf{Pnx} results are underlined.}
\label{tab:all-dev-results}
\end{table*}

\newpage
\onecolumn
\section{LLMs Results}
\label{sec:appendix-llms-res}
% https://docs.google.com/spreadsheets/d/1LPT-S8FZBSbl85r02KGMECYCsPt9POgQSjuTLpdsdWc/edit?gid=1713424700#gid=1713424700

\begin{table*}[h]
\setlength{\tabcolsep}{4.3pt}
\small
\centering
\begin{tabular}{lcc|cccc|cccc|cccc|c}
\toprule

\multicolumn{3}{c}{} & \multicolumn{4}{c}{\bf QALB-2014} & \multicolumn{4}{c}{\bf ZAEBUC} & \multicolumn{4}{c}{\bf MADAR CODA } & \textbf{ Avg.} \\

\textbf{Model} &\textbf{P-Lang} & \textbf{Shots}  & {\bf P} & {\bf R} & {\bf F\textsubscript{1}} & {\bf F\textsubscript{0.5}} & {\bf P} & {\bf R} & {\bf F\textsubscript{1}} & {\bf F\textsubscript{0.5}} & {\bf P} & {\bf R} & {\bf F\textsubscript{1}} & {\bf F\textsubscript{0.5}} & {\bf F\textsubscript{0.5}}\\
\hline

GPT-3.5-turbo & EN & 0 & 
70.6 &  54.8 &  61.7 &  66.7 &  70.8 &  70.3 &  70.5 &  70.7 &  22.8 &  17.7 &  19.9 &  21.5    & 53.0 \\
GPT-3.5-turbo & EN & 5 & 68.6 & 58.6 &  63.2 &  66.3 &  71.0 &  63.5 &  67.1 &  69.4 &  35.5 &  29.7 &  32.3 &  34.1  & \underline{56.6}  \\
GPT-3.5-turbo & AR & 0  & 70.0 &    58.5 &  63.7 &  67.3 &  68.3 &  71.3 &  69.8 &  68.9 &  24.2 &  22.7 &  23.4 &  23.9     & 53.3 \\
GPT-3.5-turbo & AR & 5  & 68.1 &    58.0 &  62.6 &  65.8 &  71.4 &  63.7 &  67.3 &  69.7 &  27.0 &  26.5 &  26.7 &  26.9     & 54.1  \\\hline

GPT-4o & EN & 0 & \textbf{82.1} &   56.4 &  66.8 &  75.2 &  80.2 &  75.5 &  77.8 &  79.2 &  28.8 &  25.5 &  27.0 &  28.1     & 60.8 \\
GPT-4o & EN & 5 & 80.7 &    65.7 &  72.4 &  \textbf{77.2} & \textbf{86.5 }& 76.8 &  \textbf{81.3} & \textbf{84.3} & \textbf{53.7} & \textbf{54.4} & \textbf{54.1} & \textbf{53.8 }   & \textbf{\underline{71.8}} \\
GPT-4o & AR & 0 & 78.9 &    62.8 &  69.9 &  75.1 &  77.4 &  \textbf{77.7} & 77.5 &  77.4 &  36.4 &  33.5 &  34.9 &  35.8   & 62.8  \\
GPT-4o & AR & 5 & 79.5 &    \textbf{66.8} & \textbf{72.6} & 76.6 &  82.6 &  75.7 &  79.0 &  81.1 &  50.1 &  48.6 &  49.4 &  49.8   & 69.2 \\\hline

Fanar & EN & 0 & 57.4 & 31.4 & 40.6 & 49.2 & 58.4 & 18.6 & 28.2 & 40.9 & 13.7 & 14.6 & 14.1 & 13.9 & 34.7 \\
Fanar & EN & 5 & 63.3 & 58.8 & 61.0 & 62.4 & 69.2 & 63.5 & 66.2 & 68.0 & 22.4 & 26.8 & 24.4 & 23.1 & 51.2 \\
Fanar & AR & 0  & 62.4 & 57.3 & 59.7 & 61.3 & 57.5 & 33.9 & 42.6 & 50.4 & 17.2 & 19.0 & 18.1 & 17.5 & 43.1  \\
Fanar & AR & 5  & 69.7 & 63.7 & 66.6 & 68.4 & 76.3 & 73.6 & 74.9 & 75.8 & 24.5 & 28.8 & 26.4 & 25.2 & \underline{56.5} \\\hline

Jais-13B-Chat & EN & 0 & 49.1 & 37.0 &  42.2 &  46.1 &  53.3 &  5.5 &   10.0 &  19.5 &  8.6 &   8.3 &   8.4 &   8.5    & 24.7  \\
Jais-13B-Chat & EN & 5 & 48.9 & 36.0 &  41.5 &  45.7 &  46.6 &  4.7 &   8.6 &   16.9 &  14.9 &  16.3 &  15.6 &  15.2   & 25.9  \\
Jais-13B-Chat & AR & 0 & 48.2 & 36.2 &  41.4 &  45.2 &  40.8 &  5.4 &   9.6 &   17.7 &  10.7 &  10.6 &  10.7 &  10.7    & 24.5  \\
Jais-13B-Chat & AR & 5 & 49.1 & 36.9 &  42.1 &  46.0 &  50.2 &  19.7 &  28.3 &  38.4 &  14.1 &  15.0 &  14.5 &  14.3    & \underline{32.9}  \\

\bottomrule

\end{tabular}
\caption{LLMs results on MSA and DA GEC on the Dev sets of QALB-2014, ZAEBUC, and MADAR CODA. P-Lang is the prompt language either in English (EN) or Arabic (AR). Best average F\textsubscript{0.5} results for each LLM are underlined; best overall results are in bold.}
\label{tab:llm-results}
\end{table*}

\section{MSA GEC Results}
\label{sec:appendix-gec-res-full-baselines}
\begin{table*}[th]
\setlength{\tabcolsep}{4pt}
% \small
\centering
\begin{tabular}{l  cccc | cccc}
\toprule

& \multicolumn{4}{c|}{\bf QALB-2014} & \multicolumn{4}{c}{\bf ZAEBUC} \\
% \midrule
& {\bf P} & {\bf R} & {\bf F\textsubscript{1}} & {\bf F\textsubscript{0.5}}\phantom{\textsuperscript{$\dagger$}} & {\bf P} & {\bf R} & {\bf F\textsubscript{1}} & {\bf F\textsubscript{0.5}}\phantom{\textsuperscript{$\dagger$}} \\
\hline
AraBART & 83.2 & 64.9 & 72.9 & 78.7\phantom{\textsuperscript{$\dagger$}}  & 87.3 & 70.6 & 78.1 & 83.4\phantom{\textsuperscript{$\dagger$}} \\
AraT5+Morph+GED\textsuperscript{43} & 83.1 & 67.9 & 74.7 & 79.6\phantom{\textsuperscript{$\dagger$}} & 85.2 & 71.2 & 77.6 & 82.0\phantom{\textsuperscript{$\dagger$}} \\
AraBART+Morph+GED\textsuperscript{13} & \underline{83.9} & 65.7 & 73.7 & 79.5\phantom{\textsuperscript{$\dagger$}} & \underline{87.6} & 73.9 & 80.2 & \underline{84.5}\phantom{\textsuperscript{$\dagger$}} \\
\hline
GPT-3.5-turbo & 68.6 & 58.6 & 63.2 & 66.3\phantom{\textsuperscript{$\dagger$}} & 71.0 & 63.5 & 67.1 & 69.4\phantom{\textsuperscript{$\dagger$}}  \\
GPT-4o & 80.7 & 65.7 & 72.4 & 77.2\phantom{\textsuperscript{$\dagger$}}  & 86.5 & \textbf{\underline{76.8}} & \textbf{\underline{81.3}} & 84.3\phantom{\textsuperscript{$\dagger$}}  \\
Fanar & 69.7 & 63.7 & 66.6 & 68.4\phantom{\textsuperscript{$\dagger$}}  & 76.3 & 73.6 & 74.9 & 75.8\phantom{\textsuperscript{$\dagger$}}  \\
Jais-13B-Chat & 49.1 & 36.9 & 42.1 & 46.0\phantom{\textsuperscript{$\dagger$}}  & 50.2 & 19.7 & 28.3 & 38.4\phantom{\textsuperscript{$\dagger$}}  \\
\hline
\textsc{sweet} & 81.8 & 68.8 & 74.7 & 78.8\phantom{\textsuperscript{$\dagger$}}  & 85.8 & 72.3 & 78.4 & 82.7\phantom{\textsuperscript{$\dagger$}}  \\
$\textsc{sweet}^{2}$& 81.9 & \textbf{\underline{70.4}} & \underline{75.7} & 79.3\phantom{\textsuperscript{$\dagger$}}  & 85.8 & 73.3 & 79.1 & 83.0\phantom{\textsuperscript{$\dagger$}}  \\
$\textsc{sweet}^{2}_{\text{NoPnx}}$ + $\textsc{sweet}^{}_{\text{Pnx}}$ & 83.7 & 68.8 & 75.6 & \underline{80.3}\textsuperscript{$\dagger$} & 86.7 & 73.9 & 79.8 & 83.8\phantom{\textsuperscript{$\dagger$}}  \\
\hline\hline
3-Ensemble & 84.9 & 68.8 & \textbf{76.0} & 81.1\phantom{\textsuperscript{$\dagger$}}  & 89.6 & 72.8 & 80.3 & 85.6\phantom{\textsuperscript{$\dagger$}}  \\
4-Ensemble & \textbf{89.1} & 61.6 & 72.8 & \textbf{81.8}\textsuperscript{$\ddagger$} & \textbf{93.3} & 68.3 & 78.9 & \textbf{86.9}\textsuperscript{$\dagger$}  \\
\bottomrule

\end{tabular}
\caption{MSA GEC results on the Dev sets of QALB-2014 and ZAEBUC. Best non-ensemble results are underlined; best overall results are in bold. $\dagger$ denotes statistical significance over the best baseline; $\ddagger$ denotes statistical significance over both the best baseline and the best non-ensemble model.}
\label{tab:gec-dev-res-all-baselines}
\vspace{-5pt}
\end{table*}

\begin{table*}[th]
\setlength{\tabcolsep}{3.6pt}
% \small
\centering
\begin{tabular}{l  cccc | cccc | cccc}
\toprule

& \multicolumn{4}{c|}{\bf QALB-2014} & \multicolumn{4}{c|}{\bf QALB-2015} & \multicolumn{4}{c}{\bf ZAEBUC} \\
% \midrule
& {\bf P} & {\bf R} & {\bf F\textsubscript{1}} & {\bf F\textsubscript{0.5}}\phantom{\textsuperscript{$\dagger$}} & {\bf P} & {\bf R} & {\bf F\textsubscript{1}} & {\bf F\textsubscript{0.5}}\phantom{\textsuperscript{$\dagger$}} & {\bf P} & {\bf R} & {\bf F\textsubscript{1}} & {\bf F\textsubscript{0.5}}\phantom{\textsuperscript{$\dagger$}} \\
\hline
AraBART & 84.0 & 64.7 & 73.1  & 79.3\phantom{\textsuperscript{$\dagger$}} & 82.0 & 71.7 & 76.5 & 79.7\phantom{\textsuperscript{$\dagger$}} & \underline{86.0} & 71.6 & 78.2 & 82.7\phantom{\textsuperscript{$\dagger$}} \\
AraBART+GED\textsuperscript{43} & 84.2 & 65.4 & 73.6 & 79.6\phantom{\textsuperscript{$\dagger$}} & 81.2 & 72.4 & 76.5 & 79.3\phantom{\textsuperscript{$\dagger$}} & 85.4 & 72.6 & 78.5 & 82.5\phantom{\textsuperscript{$\dagger$}} \\
AraBART+Morph+GED\textsuperscript{43} & 83.9 & 65.7 & 73.7 & 79.5\phantom{\textsuperscript{$\dagger$}} & \underline{82.6} & 72.1 & 77.0 & \underline{80.3}\phantom{\textsuperscript{$\dagger$}} & 85.4 & 73.7 & 79.1 & 82.7\phantom{\textsuperscript{$\dagger$}} \\
AraBART+GED\textsuperscript{13} & 84.1 & 65.0 & 73.3 & 79.4\phantom{\textsuperscript{$\dagger$}} & 81.5 & 72.7 & 76.8 & 79.5\phantom{\textsuperscript{$\dagger$}} & 85.9 & 73.4 & 79.2 & \underline{83.1}\phantom{\textsuperscript{$\dagger$}} \\
\hline
GPT-4o & 81.5 & 65.5 & 72.6 & 77.7\phantom{\textsuperscript{$\dagger$}} & 81.1 & \textbf{\underline{74.3}} & 77.5 & 79.6\phantom{\textsuperscript{$\dagger$}} & 84.4 & \textbf{\underline{75.9}} & \underline{79.9} & 82.5\phantom{\textsuperscript{$\dagger$}} \\
\hline
$\textsc{sweet}^{2}$& 82.6 & \textbf{\underline{69.5}} & \textbf{\underline{75.5}} & 79.6\phantom{\textsuperscript{$\dagger$}} & 80.0 & \textbf{\underline{74.3}} & 77.0 & 78.8\phantom{\textsuperscript{$\dagger$}} & 85.5 & 74.4 & 79.6 & 83.0\phantom{\textsuperscript{$\dagger$}} \\
$\textsc{sweet}^{2}_{\text{NoPnx}}$ + $\textsc{sweet}^{}_{\text{Pnx}}$ & \underline{84.5} & 67.7 & 75.2 & \underline{80.5}\textsuperscript{$\dagger$} & 82.2 & 73.6 & \underline{77.7} & \underline{80.3}\phantom{\textsuperscript{$\dagger$}} & 85.7 & 74.1 & 79.5 & \underline{83.1}\phantom{\textsuperscript{$\dagger$}} \\
\hline\hline
3-Ensemble & 85.7 & 67.4 & 75.4 & 81.3\phantom{\textsuperscript{$\dagger$}} & 83.7 & 73.3 & \textbf{78.1} & 81.3\phantom{\textsuperscript{$\dagger$}} & 89.7 & 73.7 & \textbf{80.9} & 85.9\phantom{\textsuperscript{$\dagger$}} \\
4-Ensemble & \textbf{89.7} & 60.2 & 72.0 & \textbf{81.7}\textsuperscript{$\ddagger$} & \textbf{88.3} & 66.7 & 76.0 & \textbf{82.9}\textsuperscript{$\ddagger$} & \textbf{93.4} & 68.9 & 79.3 & \textbf{87.2}\textsuperscript{$\ddagger$} \\
\bottomrule

\end{tabular}
\caption{MSA GEC results on the Test sets of QALB-2014, QALB-2015 (L1), and ZAEBUC. Best non-ensemble results are underlined; best overall results are in bold. $\dagger$ denotes statistical significance over the best baseline; $\ddagger$ denotes statistical significance over both the best baseline and the best non-ensemble model.}
\label{tab:gec-test-res-all-baselines}
\end{table*}

% https://docs.google.com/spreadsheets/d/1LPT-S8FZBSbl85r02KGMECYCsPt9POgQSjuTLpdsdWc/edit?gid=1373417432#gid=1373417432

\newpage
\section{DA GEC Results}
\label{sec:appendix-coda-res-full-baselines}

% \begin{table}[t]
% \setlength{\tabcolsep}{3pt}
% % \small
% \centering
% \begin{tabular}{l  lllll}
% \toprule

% % & \multicolumn{5}{c}{\bf MADAR-CODA}  \\
% % \midrule
% & {\bf P} & {\bf R} & {\bf F\textsubscript{1}} & {\bf F\textsubscript{0.5}} & {\bf WER}  \\
% \hline
% A'2024 (Seq2Seq) & 86.8 & 77.4 & 81.8 & 84.7 & 0.062 \\
% A'2024 (Seq2Seq++) & 87.6 & 79.3 & 83.3 & 85.8 & 0.057 \\\hline
% GPT-3.5-turbo & 35.5 & 29.7 & 32.3 & 34.1 & 0.324 \\
% GPT-4o & 53.7 & 54.4 & 54.1 & 53.8 & 0.189 \\
% Fanar & 16.0 & 17.1 & 16.5 & 16.2 & 0.652  \\
% Jais-13B-Chat & 14.1 & 15.0 & 14.5 & 14.3 & 0.984  \\\hline
% \textsc{sweet} & 89.1 & 75.5 & 81.7 & 86.0 & 0.067  \\
% $\textsc{sweet}^{\text{2}}$ & 87.5 & 73.5 & 79.9 & 84.3 & 0.071  \\\hline\hline
% 3-Ensemble & 91.7 & 77.4 & 83.9 & 88.4 & 0.060  \\
% + GPT-4o & 93.8 & 72.5 & 81.8 & 88.6 & 0.071  \\
% \hline

% \end{tabular}
% \caption{CODAfication results on the Dev set of MADAR-CODA. A'2024 refers to \newcite{alhafni-etal-2024-exploiting}}
% \label{tab:data-stats}
% \end{table}

\begin{table}[h]
\setlength{\tabcolsep}{4pt}
% \small
\centering
\begin{tabular}{l  cccc}
\toprule

% & \multicolumn{5}{c}{\bf MADAR-CODA}  \\
% \midrule
& {\bf P} & {\bf R} & {\bf F\textsubscript{1}} & {\bf F\textsubscript{0.5}}\phantom{\textsuperscript{$\dagger$}}  \\
\hline
AraT5 & 86.8 & 77.4 & 81.8 & 84.7\phantom{\textsuperscript{$\dagger$}} \\
AraT5+City & 87.6 & \textbf{\underline{79.3}} & \underline{83.3} & 85.8\phantom{\textsuperscript{$\dagger$}}  \\\hline
GPT-3.5-turbo & 35.5 & 29.7 & 32.3 & 34.1\phantom{\textsuperscript{$\dagger$}}  \\
GPT-4o & 53.7 & 54.4 & 54.1 & 53.8\phantom{\textsuperscript{$\dagger$}}  \\
Fanar & 24.5 & 28.8 & 26.4 & 25.2\phantom{\textsuperscript{$\dagger$}}   \\
Jais-13B-Chat & 14.1 & 15.0 & 14.5 & 14.3\phantom{\textsuperscript{$\dagger$}}   \\\hline
\textsc{sweet} & \underline{89.1} & 75.5 & 81.7 & \underline{86.0}\phantom{\textsuperscript{$\dagger$}}  \\
$\textsc{sweet}^{\text{2}}$ & 87.5 & 73.5 & 79.9 & 84.3\phantom{\textsuperscript{$\dagger$}}   \\\hline\hline
3-Ensemble & 91.7 & 77.4 & \textbf{83.9} & 88.4\phantom{\textsuperscript{$\dagger$}}  \\
4-Ensemble & \textbf{93.8} & 72.5 & 81.8 & \textbf{88.6}\textsuperscript{$\ddagger$}   \\
\bottomrule

\end{tabular}
\caption{DA GEC results on the MADAR CODA Dev set. Best non-ensemble results are underlined; best overall results are in bold. $\ddagger$ denotes statistical significance over both the best baseline and the best non-ensemble model.}
\vspace{-5pt}

\label{tab:coda-dev-res-all-baselines}
\end{table}

% \begin{table}[t]
% \setlength{\tabcolsep}{3.5pt}
% % \small
% \centering
% \begin{tabular}{l  lllll}
% \toprule

% % & \multicolumn{5}{c}{\bf MADAR-CODA}  \\
% % \midrule
% & {\bf P} & {\bf R} & {\bf F\textsubscript{1}} & {\bf F\textsubscript{0.5}} & {\bf WER}  \\
% \hline
% A'2024 (S2S) & 86.8 & 77.4 & 81.8 & 84.7 & 0.062 \\
% A'2024 (S2S++) & 87.6 & 79.3 & 83.3 & 85.8 & 0.057 \\\hline
% GPT-4o & 53.7 & 54.4 & 54.1 & 53.8 & 0.189 \\\hline
% \textsc{sweet} & 87.5 & 73.5 & 79.9 & 84.3 & 0.071  \\\hline\hline
% 3-Ensemble & 91.7 & 77.4 & 83.9 & 88.4 & 0.060  \\
% + GPT-4o & 93.8 & 72.5 & 81.8 & 88.6 & 0.071  \\
% \hline

% \end{tabular}
% \caption{CODAfication results on the Test set of MADAR-CODA. A'2024 refers to \newcite{alhafni-etal-2024-exploiting}. \textcolor{red}{The ensemble here includes the best CAMeLBERT model. Should we report test results on CAMeLBERT separately?}}
% \label{tab:data-stats}
% \end{table}

\begin{table}[h!]
\setlength{\tabcolsep}{4pt}
% \small
\centering
\begin{tabular}{l  cccc}
\toprule

% & \multicolumn{5}{c}{\bf MADAR-CODA}  \\
% \midrule
& {\bf P} & {\bf R} & {\bf F\textsubscript{1}} & {\bf F\textsubscript{0.5}}\phantom{\textsuperscript{$\dagger$}}   \\
\hline
AraT5 & 87.3 & 78.0 & 82.4 & 85.2\phantom{\textsuperscript{$\dagger$}}  \\
AraT5+DA Phrase & 88.4 & \textbf{\underline{79.0}} & \underline{83.4} & 86.3\phantom{\textsuperscript{$\dagger$}}  \\\hline
GPT-4o & 56.1 & 54.8 & 55.5 & 55.9\phantom{\textsuperscript{$\dagger$}} \\\hline
\textsc{sweet} & \underline{89.4} & 76.6 & 82.5 & \underline{86.5}\phantom{\textsuperscript{$\dagger$}}   \\\hline\hline
3-Ensemble & 92.2 & 77.7 & \textbf{84.3} & \textbf{88.9}\phantom{\textsuperscript{$\dagger$}}  \\
4-Ensemble & \textbf{94.0} & 72.9 & 82.1 & 88.8\textsuperscript{$\ddagger$}  \\
\bottomrule

\end{tabular}
\caption{DA GEC results on the Test set of MADAR CODA. Best non-ensemble results are underlined; best overall results are in bold. $\ddagger$ denotes statistical significance over both the best baseline and the best non-ensemble model.}
% \textcolor{red}{The ensemble here includes the best CAMeLBERT model. Should we report test results on CAMeLBERT separately?}}
\label{tab:coda-test-res-all-baselines}
\end{table}

\clearpage
\onecolumn

\section{Error Type Statistics}
\label{sec:appendix-error-type-stats}
\setlength\dashlinedash{1pt}
\setlength\dashlinegap{1.5pt}
\setlength\arrayrulewidth{0.3pt}

\begin{table*}[h]
\centering
\begin{tabular}{llll|lll|lll}
    \toprule
        \multicolumn{1}{l}{} & \multicolumn{3}{c}{\textbf{QALB-2014}} & \multicolumn{3}{c}{\textbf{ZAEBUC}} & \multicolumn{3}{c}{\textbf{MADAR CODA}} \\
        \multicolumn{1}{l}{} & \textbf{Train} & \textbf{Dev} & \textbf{Test} & \textbf{Train} & \textbf{Dev} & \textbf{Test} & \textbf{Train} & \textbf{Dev} & \textbf{Test} \\\hline

        Delete  &  6,442 & 346 & 540 & 305 & 64 & 66 & 35 & 0 & 1 \\
        Merge-B &  15,063 & 797 & 795 & 849 & 180 & 133 & 404 & 102 & 95 \\
        Merge-I &  15,296 & 812 & 807 & 851 & 180 & 133 & 429 & 109 & 103 \\
        M       &  1,466 & 69 & 63 & 137 & 32 & 28 & 159 & 42 & 56 \\
        M+O     &  243 & 17 & 15 & 7 & 1 & 8 & 0 & 0 & 0 \\
        O       &  141,752 & 7,380 & 6,976 & 3,203 & 695 & 792 & 5,604 & 1,179 & 1,193 \\
        O+X     &  323 & 24 & 18 & 20 & 0 & 4 & 0 & 0 & 0 \\
        P       &  11,379 & 598 & 687 & 237 & 51 & 36 & 18 & 10 & 9 \\
        S       &  5,436 & 247 & 252 & 169 & 36 & 51 & 408 & 77 & 64 \\
        X       &  13,592 & 809 & 743 & 528 & 110 & 113 & 911 & 166 & 151 \\
        Split   &  7,828 & 432 & 399 & 49 & 10 & 10 & 279 & 58 & 69 \\
        UNK     &  6,835 & 331 & 300 & 361 & 78 & 61 & 2,235 & 430 & 649 \\
        C       &  795,510 & 41,875 & 39,690 & 18,411 & 3,839 & 3,683 & 29,285 & 6,627 & 6,456 \\\hline
                &  1,021,165 & 53,737 & 51,285 & 25,127 & 5,276 & 5,118 & 39,767 & 8,800 & 8,846 \\

    \bottomrule
\end{tabular}
\caption{Distribution of error types in QALB-2014, ZAEBUC, and MADAR CODA. UNK refers to unknown error types; C refers to correct words.} 
\label{tab:error-type-stats}
\end{table*}

\clearpage
\onecolumn

\section{Prompts}
\label{sec:appendix-prompts}
\begin{figure*}[h]
\centering
\includegraphics[width=\textwidth]{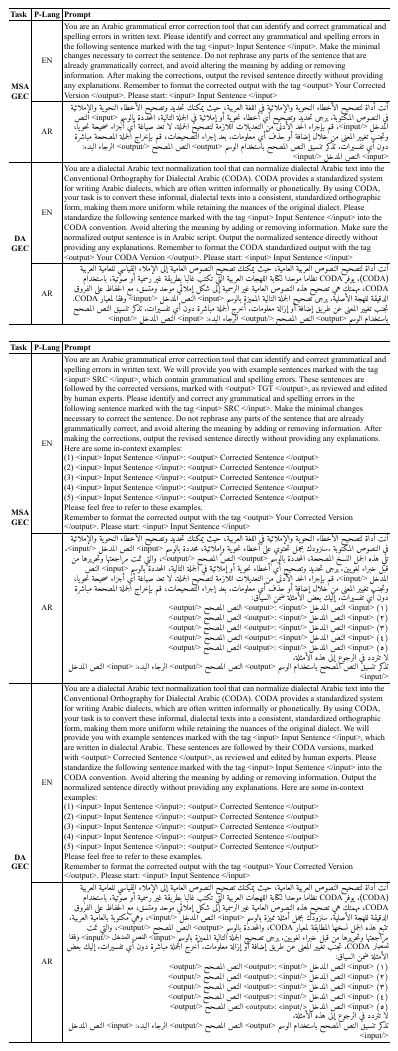}
\caption{0-shot prompts used to evaluate LLMs performance on MSA and DA GEC. P-Lang is the prompt language either in English (EN) or Arabic (AR).}
\label{fig:zero-shot-prompts}
\end{figure*}

\clearpage
\onecolumn

\begin{figure*}[h]
\centering
\includegraphics[width=0.94\textwidth]{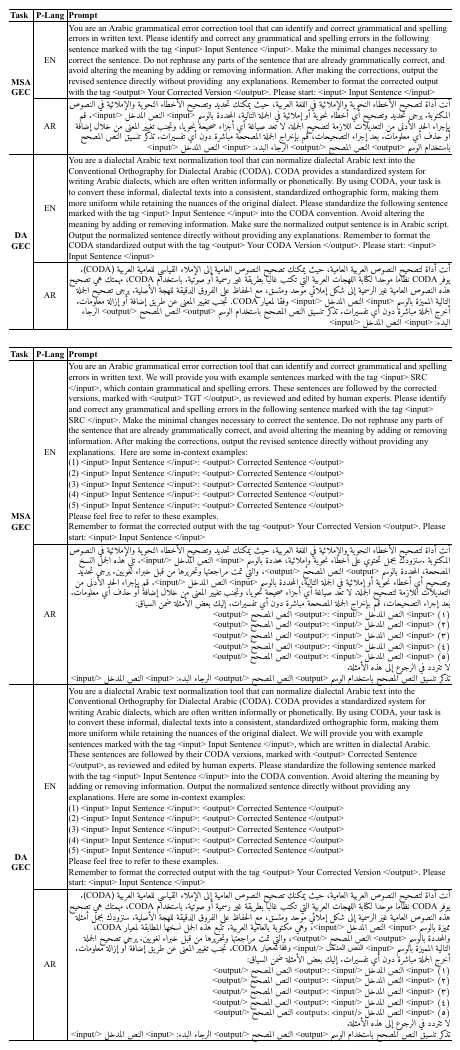}
\caption{5-shot prompts used to evaluate LLMs performance on MSA and DA GEC. P-Lang is the prompt language either in English (EN) or Arabic (AR).}
\label{fig:few-shot-prompts}
\end{figure*}

\end{document}